\documentclass[letterpaper]{article} 
\usepackage{aaai2026}  
\usepackage{times}  
\usepackage{helvet}  
\usepackage{courier}  
\usepackage[hyphens]{url}  
\usepackage{graphicx} 
\usepackage{natbib}
\usepackage{caption}
\usepackage{algorithm}
\usepackage{amsmath}
\usepackage{amsfonts}
\usepackage{array}
\usepackage{algpseudocode}
\usepackage{booktabs}
\usepackage{multirow}
\usepackage{subcaption}
\usepackage{makecell}
\usepackage{newfloat}
\usepackage{listings}

\usepackage{newfloat}
\usepackage{listings}
\DeclareCaptionStyle{ruled}{labelfont=normalfont,labelsep=colon,strut=off} 
\floatstyle{ruled}
\newfloat{listing}{tb}{lst}{}
\floatname{listing}{Listing}
\title{Advanced Black-Box Tuning of Large Language Models with Limited API Calls}
\author {
    Zhikang Xie\textsuperscript{\rm 1}, 
    Weilin Wan\textsuperscript{\rm 1}, 
    Peizhu Gong\textsuperscript{\rm 1}\thanks{Corresponding Author.}, 
    Weizhong Zhang\textsuperscript{\rm 2},
    Cheng Jin\textsuperscript{\rm 1,3}\footnotemark[1]
}
\affiliations {
    \textsuperscript{\rm 1} College of Computer Science and Artificial Intelligence, Fudan University \\
    \textsuperscript{\rm 2} School of Data Science, Fudan University \\
    \textsuperscript{\rm 3} Shanghai Key Laboratory of Intelligent Information Processing \\
    \{22307110187, wlwan23\}@m.fudan.edu.cn, 
    \{pzgong, weizhongzhang, jc\}@fudan.edu.cn
}

\usepackage{bibentry}

\begin{document}

\maketitle

\begin{abstract}
Black-box tuning is an emerging paradigm for adapting large language models (LLMs) to better achieve desired behaviors, particularly when direct access to model parameters is unavailable. Current strategies, however, often present a dilemma of suboptimal extremes: either separately train a small proxy model and then use it to shift the predictions of the foundation model, offering notable efficiency but often yielding limited improvement; or making API calls in each tuning iteration to the foundation model, which entails prohibitive computational costs. In this paper, we argue that a more reasonable way for black-box tuning is to train the proxy model with limited API calls. The underlying intuition is based on two key observations: first, the training samples may exhibit correlations and redundancies, suggesting that the foundation model’s predictions can be estimated from previous calls; second, foundation models frequently demonstrate low accuracy on downstream tasks. Therefore, we propose a novel advanced black-box tuning method for LLMs with limited API calls. Our core strategy involves training a Gaussian Process (GP) surrogate model with “LogitMap Pairs" derived from querying the foundation model on a minimal but highly informative training subset. This surrogate can approximate the outputs of the foundation model to guide the training of the proxy model, thereby effectively reducing the need for direct queries to the foundation model. Extensive experiments verify that our approach elevates pre-trained language model accuracy from \textbf{55.92\%} to \textbf{86.85\%}, reducing the frequency of API queries to merely \textbf{1.38\%}. This significantly outperforms offline approaches that operate entirely without API access. Notably, our method also achieves comparable or superior accuracy to query-intensive approaches, while significantly reducing API costs. This offers a robust and high-efficiency paradigm for language model adaptation.
\end{abstract}

\begin{links}
    \link{Code}{https://github.com/kurumi8686/EfficientBBT}
\end{links}

\section{Introduction}
\label{introduction}

Large Language Models (LLMs) have demonstrated remarkable capabilities in recent years. Adapting them to specific downstream tasks or aligning them with desired behaviors is essential for unlocking their full potential in real-world applications. Gradient-based methods, such as Adapter modules \cite{houlsby2019parameter} and LoRA \cite{hu2022lora}, are widely recognized as standard parameter-efficient fine-tuning techniques for LLMs. These methods adapt the pre-trained model to new tasks by tuning only a small subset of the model's parameters, and they have consistently achieved promising results in the literature. However, these methods require full access to model parameters, which is not feasible for many state-of-the-art LLMs, such as GPT-4 \cite{achiam2023gpt} and Gemini \cite{team2023gemini}.

Black-box tuning is an emerging paradigm for adapting LLMs without direct parameter access. However, current strategies often involve a challenging trade-off. Specifically, \textbf{offline methods}~\cite{liu2024tuning} train a smaller proxy model independently and use it to adjust the black-box model's outputs during inference. While these methods are efficient, their performance is limited because the proxy model does not have direct access to the foundation model’s internal knowledge during training. In contrast, \textbf{online methods}, such as Consistent Proxy Tuning (CPT) \cite{he2024cpt}, integrate the black-box model into the proxy’s training loop via iterative API calls. This approach improves alignment and performance but incurs significant computational and monetary costs. Consequently, practitioners face a dilemma: either sacrifice performance for efficiency or accept substantial costs for better adaptation.

\begin{figure*}[!t]
  \centering 
  \includegraphics[width=0.99\textwidth]{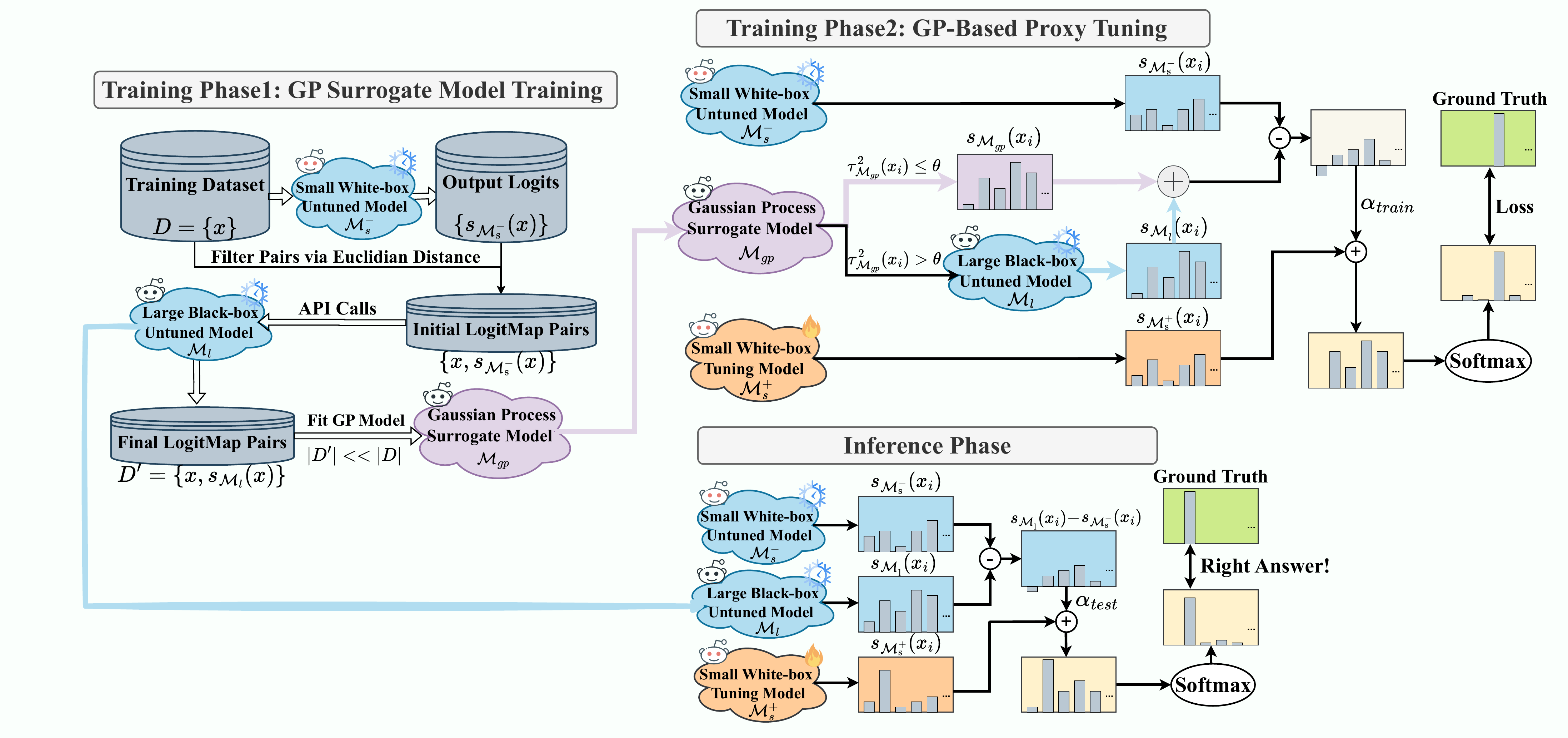}
  \caption{Overview of our proposed algorithmic framework.
  \textbf{Training Phase 1: GP Surrogate Model Training.} A Gaussian Process model $\mathcal{M}_{gp}$ is trained on a filtered subset of data to approximate the mapping between input embeddings and the output logits of the large black-box model $\mathcal{M}_{l}$. 
  \textbf{Training Phase 2: GP-based Proxy Tuning.} The trained $\mathcal{M}_{gp}$ guides the fine-tuning of a small white-box proxy model $\mathcal{M}^+_s$, effectively incorporate knowledge from $\mathcal{M}_l$ into $\mathcal{M}^+_s$ via standard supervised training. 
  \textbf{Inference Phase.} The final predictions are obtained by combining the outputs from the tuned proxy model \(\mathcal{M}^+_s\) with an ensemble of the original proxy model \(\mathcal{M}^-_s\) and the large black-box model \(\mathcal{M}_l\).}
  \label{fig:method}
\end{figure*}

In this paper, we argue that a more reasonable and resource-efficient way for black-box tuning is to train a proxy model with a strictly limited budget of these costly API calls. The underlying intuition for this approach is rooted in two key observations. Firstly, the training samples may exhibit inherent correlations and redundancies. This suggests that the foundation model’s predictions for new, unseen inputs can often be effectively estimated or inferred from its responses to a smaller subset of previous calls. Secondly, even powerful foundation models do not always achieve perfect accuracy across all instances within the training data. Consequently, we posit that a well-informed approximation, rather than exhaustive querying of the foundation model, can still provide effective and comparable supervision for training a high-performing proxy model.

Therefore, we propose a novel advanced black-box tuning method specifically designed for LLMs operating under the limited API calls. 
Our core strategy involves leveraging a Gaussian Process (GP) surrogate \cite{williams2006gaussian} to approximate the outputs of the foundation model to guide the training of the proxy model. 
As illustrated in Figure~\ref{fig:method}, the central idea is to train this GP surrogate model on a small yet highly informative subset of the training data. We refer to these data points as “LogitMap Pairs", which consist of input embeddings and their corresponding output logits obtained from the foundation model.
Once trained, the GP effectively approximates the foundation’s predictive behavior, thereby enabling logit-level supervision for a lightweight proxy model and highly reducing the need for expensive black-box queries.

GPs, as non-parametric Bayesian models, are exceptionally well-suited for approximating complex functions and have been shown to emulate deep neural networks under certain conditions \cite{damianou2013deep, lee2017deep}. A crucial advantage of GPs is their probabilistic nature, which allows for robust uncertainty estimation. This capability can help quantify the reliability of the surrogate's predictions and further inform strategic training decisions for the proxy model. We leverage this property to enhance the proxy model training process. Specifically, when the GP surrogate yields a prediction with high associated uncertainty (e.g., a large variance $ \tau^2 $ exceeding a pre-defined threshold $ \theta $), we deem its prediction potentially unreliable. In such instances, our method falls back on invoking the black-box target model to obtain the true output.
This adaptive mechanism ensures that primarily high-confidence surrogate predictions are utilized for training, thereby reducing the propagation of noise and improving the overall robustness and capabilities of the proxy model.

Extensive experiments across multiple NLP benchmarks demonstrate the effectiveness and scalability of our approach. It boosts the average accuracy of pre-trained language models from \textbf{55.92\%} to \textbf{86.85\%}, while reducing API query usage to just \textbf{1.38\%} of that required by query-intensive methods such as CPT \cite{he2024cpt}. Remarkably, despite this drastic reduction, our method achieves comparable or superior accuracy to these online methods, and significantly outperforms fully offline baselines. This highlights our approach as a robust and highly efficient paradigm for adapting LLMs in black-box settings.

Our contributions can be summarized as follows:
\begin{enumerate}
    \item We propose a novel black-box tuning method that employs a GP surrogate to approximate foundation model outputs, enabling efficient proxy training with minimal API queries. As far as we have studied, our method improves from \textbf{55.92\%} to \textbf{86.85\%}, achieving \textbf{state-of-the-art} results and demonstrating its superior effectiveness.
    \item We propose an effective data selection method for GP model training, requiring only \textbf{1.38\%} training data.
    \item Extensive experiments show that our method drastically reduces API usage compared to previous online approaches, while achieving better performance. This demonstrates a practical and cost-efficient paradigm for black-box tuning in real-world scenarios.
\end{enumerate}

\section{Related Works}
\label{related_works}

\paragraph{Efficient Fine-tuning.}
The substantial cost of fully fine-tuning large models \cite{roziere2023code, groeneveld2024olmo} has driven the development of parametric-efficient fine-tuning (PEFT) techniques \cite{he2021towards, lialin2023scaling}. 
PEFT methods adapt models by modifying only a small parameter subset, aiming to preserve pre-trained knowledge while reducing resource demands. Common strategies involve inserting lightweight modules (e.g., Adapters \cite{houlsby2019parameter}, Compacter \cite{karimi2021compacter}), optimizing continuous prompts or prefixes (e.g., Prompt Tuning \cite{lester2021power}, Prefix Tuning \cite{li2021prefix}, P-Tuning v2 \cite{liu2021p}), or adjusting internal model parameters through approaches like low-rank updates (LoRA \cite{hu2022lora}, QLoRA \cite{dettmers2023qlora}), selective tuning (e.g., BitFit \cite{zaken2021bitfit}), or learned activation scaling ($(\mathrm{IA})^3$ \cite{liu2022few}). 
Despite their resource efficiency, these approaches typically require internal model access (weights and gradients), restricting their use in black-box scenarios.

\paragraph{Black-box Fine-tuning.} 
Adapting LLMs without parameter access (i.e., in black-box settings) requires specialized fine-tuning techniques. 
Gradient-free optimization offers one approach, exemplified by Black-Box Tuning (BBT) \cite{sun2022black}, which optimizes input prompts by evaluating model outputs without gradient information. A dominant alternative involves employing smaller auxiliary models. Proxy-based methods, such as Proxy Tuning (PT) \cite{liu2024tuning} and Consistent Proxy Tuning (CPT) \cite{he2024cpt}, train an accessible white-box proxy and transfer task-specific knowledge by using differential signals from the proxy to guide the black-box model at inference. Other strategies train surrogate models to post-process or align the black-box outputs directly, using methods like sequence-to-sequence aligners \cite{ji2024aligner} or adapting output probabilities \cite{ormazabal2023comblm, lu2023inference}. A key challenge across many methods relying on auxiliary models, especially CPT, is the potentially high cost associated with frequent API queries to the black-box LLM needed for training the auxiliary component.

\paragraph{Logit Arithmetic.}
Involving techniques that directly manipulate pre-softmax logits, often by aggregating signals from multiple sources or model states, is essentially an application of ensemble learning principles \cite{dong2020survey}. For example, logit manipulation facilitates domain adaptation through ensembling logits from distinct models \cite{dou2019domain}. In controllable generation, approaches include subtracting anti-expert logits (DExperts \cite{liu2021dexperts}) and contrasting expert versus amateur model logits (CD \cite{li2022contrastive}). More recently, the principle has been extended to intra-model comparisons, where different layers are contrasted to enhance factuality (DoLa \cite{chuang2023dola}) or guide decoding via auto-contrastive objectives \cite{gera2023benefits}. The effectiveness and flexibility of logit ensembling motivate our exploration of logit-based adjustments as an efficient mechanism for black-box tuning.

\paragraph{Gaussian Process Models.}
Gaussian Processes (GPs) are non-parametric Bayesian methods well-suited for modeling complex functions and quantifying uncertainty \cite{williams2006gaussian}. Key advances include sophisticated covariance functions, such as additive kernels for interpretable decomposition \cite{durrande2011additive} and multiple kernel learning for integrating diverse properties \cite{gonen2011multiple}. 
GPs have also been integrated into more complex probabilistic frameworks to capture complex data structures. Notable examples include Warped GPs \cite{snelson2003warped}, which transform the output space to model non-Gaussian likelihoods, and Gaussian Process Regression Networks \cite{wilson2011gaussian}, which compose multiple GPs into deeper hierarchical models. The broad applicability of GPs across various natural language processing (NLP) tasks \cite{cohn2014gaussian} further highlights their versatility. These advancements demonstrate the flexibility and modeling capabilities of Gaussian Processes, motivating our adoption of GP-based models as efficient surrogates.

\section{Methodology}
\label{methodology}

This section outlines our GP-based approach for efficient black-box tuning of large language models. Central to our method is a GP surrogate that approximates target models behavior, enabling high-quality adaptation with significantly fewer direct queries. 
We first provide a brief overview of existing proxy-based approaches, followed by a comprehensive and detailed description of our advanced framework, highlighting its key innovations and practical benefits.

\subsection{Basics on Existing Proxy-based Methods}

\subsubsection{Proxy-Tuning}
Proxy-Tuning (PT) \cite{liu2024tuning} adapts pre-trained LLMs at decoding time without access to their internal parameters, ideal for black-box or computationally constrained scenarios. It employs a small white-box proxy model, with a tuned version $\mathcal{M}^+_s$ and an untuned version $\mathcal{M}^-_s$. PT adjusts the logits of the large black-box model $\mathcal{M}_{l}$ by adding the logit difference from the small proxy models:
\begin{equation}
s_{\textit{pt}}(x) = s_{\mathcal{M}_{l}}(x) + \left( s_{\mathcal{M}^+_s}(x) - s_{\mathcal{M}^-_s}(x) \right),
\end{equation}
where \(s_{\mathcal{M}}(x)\) are logits from model \(\mathcal{M}\). Originally, \(\mathcal{M}^+_s\) (parameters \({\theta^+_s}\)) is trained independently on a task-specific dataset \(D = \{(x,y)\}\) to minimize a loss \(L\):
\begin{equation}
\label{eq:2}
{\theta^+_s} = \arg\min_{{\theta^+_s}} \mathbb{E}_{(x,y) \sim D} \big[ L(\mathcal{M}^+_s(x; {\theta^+_s}), y) \big],
\end{equation}
this independent training of \(\mathcal{M}^+_s\), however, overlooks its interaction with \(\mathcal{M}_{l}\) and \(\mathcal{M}^-_s\) during inference, potentially limiting performance.

\subsubsection{Consistent Proxy Tuning}
Consistent Proxy Tuning (CPT) \cite{he2024cpt} refines PT by aligning the training objective of the small proxy model \(\mathcal{M}^+_s\) with its actual usage during inference. This consistency is achieved by incorporating the influence of \(\mathcal{M}_{l}\) and \(\mathcal{M}^-_s\) into the training loss for \(\mathcal{M}^+_s\). The training objective of CPT is:
\begin{equation}
\label{eq:cpt}
\begin{split}
    {\theta^+_s} & = \arg\min_{\theta^+_s} \mathbb{E}_{(x,y) \sim D} \Big[ L \Big( \mathcal{M}^+_s(x; {\theta^+_s}) \\
    & + \alpha_{\textit{train}} \big( \mathcal{M}_{l}(x; {\theta_l}) - \mathcal{M}^-_s(x; {\theta^-_s}) \big), y \Big) \Big].
\end{split}
\end{equation}

During this training, parameters \({\theta_l}\) of \(\mathcal{M}_l\) and \({\theta^-_s}\) of \(\mathcal{M}^-_s\) are {frozen}, only \({\theta^+_s}\) are {fine-tuned}. While CPT demonstrates improved performance due to this consistent objective, a significant practical drawback arises: optimizing \(\mathcal{M}^+_s\) via Equation \ref{eq:cpt} necessitates frequent queries to the large black-box model \(\mathcal{M}_l\), incurring substantial computational costs and API call limitations.

\subsection{Proposed Method}
To mitigate the high API call dependency of CPT, we introduce a GP model as a data-efficient surrogate for the large black-box model \(\mathcal{M}_l\). The core idea is to pre-train a GP to approximate the logit outputs of \(\mathcal{M}_l\) on the task-specific dataset \(D\), and then use this GP surrogate during the CPT training of the small proxy model \(\mathcal{M}^+_s\).

\subsubsection{Gaussian Process Modeling of \(\mathcal{M}_l\) Logits}
GPs are non-parametric Bayesian models adept at function approximation from limited data. 
Formally, a GP defines a prior distribution over \(f(x) \sim \mathcal{GP}(m(x), k(x, x'))\), where \(m(x)\) is the mean function and \(k(x, x')\) is the kernel function. The kernel encodes prior beliefs about the function's properties, such as smoothness, by defining the covariance between function values at different input points \(x\) and \(x'\).

Given a training dataset \(D' = \{ (x_j, \mathbf{s}_j) \}_{j=1}^M\), where \(\mathbf{s}_j = s_{\mathcal{M}_l}(x_j) \in \mathbb{R}^V\) are the observed target black-box model output logits, the GP conditions on this data to form a posterior distribution. For a new input \(x_*\), the predictive distribution for the logits \(s_{\mathcal{M}_l}(x_*)\) is also Gaussian. A common and effective strategy for handling such multi-dimensional outputs, which we adopt in our implementation, is to model each of the \(V\) logit dimensions independently. This independent modeling assumption simplifies computation and often yields strong empirical results. 

The predictive mean for the \(v\)-th logit dimension, \(\hat{s}_{\mathcal{GP},v}(x_*)\), which serves as our approximation \(s_{\mathcal{GP},v}(x_*)\), is given by formula 2.25 and 2.27 in page 17 of \cite{williams2006gaussian}:
\begin{equation} 
\label{eq:gp_prediction} 
\hat{s}_{\mathcal{GP},v}(x_*) = \mathbf{k}(x_*, X_{D'})^T (K_{D'D'} + \sigma_{n,v}^2 I)^{-1} \mathbf{s}_{D',v},
\end{equation}
where the terms are defined as follows:
\begin{itemize}
    \item \(X_{D'} = \{x_j\}_{j=1}^M\) represents the set of \(M\) training data.
    \item \(\mathbf{k}(x_*, X_{D'})\) is a vector in $\mathbb{R}^M$  denoting the  covariances between the new input \(x_*\) and each training input \(x_j \in X_{D'}\), with its \(j\)-th element being \(k(x_*, x_j)\).
    \item \(K_{D'D'}\) is the \(M \times M\) covariance matrix computed from the training inputs, where each entry \((K_{D'D'})_{ij} = k(x_i, x_j)\) is the kernel evaluation between \(x_i, x_j \in X_{D'}\).
    \item \(\sigma_{n,v}^2\) is the noise variance hyperparameter for the \(v\)-th logit dimension, accounting for potential observation noise or model misspecification.
    \item \(I\) is the \(M \times M\) identity matrix.
    \item \(\mathbf{s}_{D',v}\) is a vector in $\mathbb{R}^M$ containing the observed values of the \(v\)-th logit dimension from the training set \(D'\).
\end{itemize}
This framework, by applying independent GPs to each logit dimension, allows us to construct a composite multi-output GP model to predict the full logit vector \(s_{\mathcal{M}_l}(x)\).

\subsubsection{Data Acquisition for GP via Selective Sampling}
\label{sample-strategies}

A critical aspect is to train the GP effectively with minimal queries to \(\mathcal{M}_l\). Instead of querying \(\mathcal{M}_l\) for all \(x \in D\), we construct a small but highly informative subset \(D' = \{ (x_j, s_{\mathcal{M}_l}(x_j)) \}_{j=1}^M\), where \(M \ll |D|\) (Typically, $D'$ is approximately 1\% the size of $D$). Then, \(D'\) is used by a filtering algorithm (detailed in Algorithm \ref{algo}) that aims to maximize diversity and representativeness.

\begin{algorithm}
\caption{Algorithm for GP Training Set Construction}
\label{algo}
\begin{algorithmic}[1]
    \Require Dataset \(D\); proxy model \(\mathcal{M}^-_s\); foundation model \(\mathcal{M}_l\); thresholds \(\tau_{\mathrm{in}}, \tau_{\mathrm{out}}\)
    \Ensure GP training set \(D'\)
    \State Initialize \(D_{\mathrm{cand}}\!=\!\emptyset\), \(D'\!=\!\emptyset\)
    \State For all \(x \in D\), compute \(\mathbf{v}_x = \mathrm{embedding}(x)\) as input, compute \(s_x = \mathcal{M}^-_s(x)\) as output
    \If{\(D\neq\emptyset\)}
        \State Seed \(D_{\mathrm{cand}}\) with \((x_1, s_{x_1})\)
    \EndIf
    \For{each \(x\in D\setminus\{x_1\}\)}
        \State \(\mathrm{diverse}\gets \text{true}\)
        \For{each \((x_k,s_k)\in D_{\mathrm{cand}}\)}
            \If{\(\|\mathbf{v}_x-\mathbf{v}_{x_k}\|\le\tau_{\mathrm{in}}\)\,\(\lor\)\,\(\|s_x-s_k\|\le\tau_{\mathrm{out}}\)}
                \State \(\mathrm{diverse}\gets \text{false}\); \textbf{break}
            \EndIf
        \EndFor
        \If{\(\mathrm{diverse}\)}
            \State Add \((x,s_x)\) to \(D_{\mathrm{cand}}\)
        \EndIf
    \EndFor
    \For{each \((x,s_x)\in D_{\mathrm{cand}}\)}
        \State Query \(\mathcal{M}_l\) for \(s'_x\); add \((x,s'_x)\) to \(D'\)
    \EndFor
    \State \Return \(D'\)
\end{algorithmic}
\end{algorithm}

This filtering process primarily uses the input vector representations $\mathbf{v}_x$ and the output logits from the frozen small model \(\mathcal{M}^-_s\). This ensures that the selection process itself is computationally inexpensive. Only once an input \(x\) is selected through this filtering, do we query the black-box large model \(\mathcal{M}_l\) to obtain its true output logits \(s_{\mathcal{M}_l}(x)\). These \((x, s_{\mathcal{M}_l}(x))\) then constitute the training set \(D'\) (i.e., LogitMap Pairs) for our GP model \(\mathcal{M}_{gp}\).

As part of this filtering strategy, we experimented with several rule-based approaches to quantify the difference. Specifically, we evaluated Manhattan distance, Euclidean distance, and cosine similarity. Our results indicate that all three metrics can perform comparably, assuming appropriate input-output thresholds are set (details in \ref{distance}). 
Among them, Euclidean distance emerged as the most consistently effective and computationally simple choice, and is therefore adopted in our final approach.

The Euclidean distances used for filtering are:
\begin{itemize}
    \item Input distance: \( d_{\text{input}}(x, x') = \| \mathbf{v}_x - \mathbf{v}_{x'} \|_2 \).
    \item Output distance: \( d_{\text{output}}(x, x') = \| s_{\mathcal{M}^-_s}(x) - s_{\mathcal{M}^-_s}(x') \|_2 \).
\end{itemize}

If two data points have highly similar input representations and their outputs (as predicted by the inexpensive small proxy model) are also similar, they likely provide redundant information for training the GP model. By filtering based on these cheaper-to-obtain metrics, we ensure \( D' \) is compact yet rich in information, capturing diverse aspects of the input space and the proxy's initial assessment of output variations. This curated selection allows the GP to generalize effectively from fewer actual \(\mathcal{M}_l\) queries.

\subsubsection{GP-Enhanced Proxy Training}
With the trained GP model $\mathcal{M}_{gp}$ providing approximation for the outputs of the large foundation model $\mathcal{M}_l$, we introduce an uncertainty-aware training objective for the small proxy model $\mathcal{M}^+_s$. The objective is defined as:
\begin{equation}
\label{eq:gp_gated}
\begin{split}
{\theta^+_s} ={}& \arg\min_{{\theta^+_s}} \mathbb{E}_{(x,y) \sim D} \Big[ L \Big(
\mathcal{M}^+_s(x; {\theta^+_s}) \\
& + \alpha_{\textit{train}} \big( S_{gate}(x) - \mathcal{M}^-_s(x; {\theta^-_s}) \big),\ y \Big) \Big],
\end{split}
\end{equation}
where the gated supervision $S_{gate}(x)$ is determined by:
\begin{equation}
\label{eq:gated_signal_definition}
\begin{split}
S_{gate}(x) ={}& \mathcal{M}_{gp}(x; {\theta_{gp}}) \cdot \mathbf{1}_{\tau^2_{\mathcal{M}_{gp}}(x) \le \theta} \\
& + \mathcal{M}_l(x; {\theta_{l}}) \cdot \mathbf{1}_{\tau^2_{\mathcal{M}_{gp}}(x) > \theta},
\end{split}
\end{equation}
where, $\tau^2_{\mathcal{M}_{gp}}(x)$ is the predictive variance of the GP model $\mathcal{M}_{gp}(x; {\theta_{gp}})$ for input $x$, and $\theta$ is the pre-defined variance threshold. 
The term $\mathbf{1}$ denotes the indicator function, where $\mathbf{1}_{\text{condition}}$ is 1 if the condition is true, and 0 otherwise.

During training, the parameters of the GP surrogate $\mathcal{M}_{gp}(\cdot; {\theta_{gp}})$ and the untuned proxy model $\mathcal{M}^-_s(\cdot; {\theta^-_s})$ are kept {frozen}, while only the trainable proxy $\mathcal{M}^+_s(\cdot; {\theta^+_s})$ is {fine-tuned}. To ensure robust supervision, we introduce a gating mechanism based on the GP’s predictive uncertainty. For each input $x$, if the GP variance $\tau^2_{\mathcal{M}_{gp}}(x) \le \theta$, its prediction is used as the guidance signal $S_{\text{gate}}(x)$; otherwise, we query the target black-box model $\mathcal{M}_l(x)$ to obtain a reliable label.

\subsubsection{Inference}
At inference time, our procedure closely follows that of PT / CPT. The adjusted logits for a new input \(x\) are computed as:
\begin{equation}
\label{eq:inference}
s_{\text{final}}(x) = s_{\mathcal{M}^+_s}(x) + \alpha_{\textit{test}} \left( s_{\mathcal{M}_l}(x) - s_{\mathcal{M}^-_s}(x) \right),
\end{equation}
where \(s_{\mathcal{M}_l}(x)\) denotes the output of the target black-box large model. Typically, we set \(\alpha_{\textit{test}} = \alpha_{\textit{train}}\). 
The final prediction is then obtained by applying the softmax function to \(s_{\text{final}}(x)\).

\section{Experiments}
\label{experiments}

\subsection{Experimental Setup}

\begin{table*}[t]
  \centering
  \renewcommand{\arraystretch}{1.5}
  \resizebox{\linewidth}{!}{
  \begin{tabular}{lccccccccccccc}
    \toprule
    \multirow{2}{*}{Model \& Method} & \multicolumn{12}{c}{\textbf{Accuracy (\%) $\uparrow$}} & \multirow{2}{*}{Avg.API $\downarrow$} \\
    \cmidrule(lr){2-13}
        & AG-News & CoLA  & CoPA  & SST-2 & ARC-C & Cs-QA & OB-QA & MNLI  & QNLI  & RTE   & QQP   & Avg. & \\
    \midrule

    \multicolumn{14}{@{}l}{\textit{Llama2-7B / Mistral-7B-v0.1}} \\
    \multicolumn{14}{@{}l}{\textit{Qwen3-8B / DeepSeek-R1-14B}} \\[1ex]
    Pretrain              
    & \makecell{52.03 / 85.33 \\ 85.86 / 87.39}  
    & \makecell{69.22 / 72.48 \\ 83.22 / 69.13} 
    & \makecell{65.40 / 87.20 \\ 93.60 / 95.00} 
    & \makecell{51.72 / 87.16 \\ 92.09 / 92.32} 
    & \makecell{40.47 / 68.56 \\ 86.62 / 88.29}
    & \makecell{26.04 / 64.21 \\ 78.13 / 77.31} 
    & \makecell{32.40 / 66.60 \\ 83.20 / 84.00} 
    & \makecell{35.74 / 36.88 \\ 84.20 / 65.79} 
    & \makecell{50.45 / 74.48 \\ 85.54 / 49.70} 
    & \makecell{54.87 / 72.56 \\ 85.56 / 52.71} 
    & \makecell{35.89 / 78.71 \\ 83.38 / 53.65} 
    & \makecell{46.75 / 72.20 \\ 85.58 / 74.12} & - \\[2ex]
    LoRA-Tune             
    & \makecell{93.59 / 93.57 \\ 90.21 / 91.30}  
    & \makecell{84.28 / 82.84 \\ 84.28 / 69.61} 
    & \makecell{81.60 / 91.20 \\ 96.20 / 86.00} 
    & \makecell{95.76 / 95.18 \\ 95.87 / 91.28} 
    & \makecell{48.83 / 70.57 \\ 90.30 / 89.63} 
    & \makecell{75.92 / 79.28 \\ 80.02 / 79.77} 
    & \makecell{75.60 / 83.60 \\ 88.40 / 90.60} 
    & \makecell{90.78 / 88.63 \\ 89.56 / 84.77} 
    & \makecell{88.36 / 88.80 \\ 88.96 / 88.25} 
    & \makecell{83.03 / 81.59 \\ 81.23 / 79.42} 
    & \makecell{89.89 / 88.10 \\ 90.34 / 90.57} 
    & \makecell{82.51 / 85.76 \\ 88.67 / 85.56} & - \\[2ex]
    Full Fine-tune        
    & \makecell{94.76 / 94.58 \\ 92.57 / 92.80}  
    & \makecell{84.85 / 84.28 \\ 84.95 / 81.59} 
    & \makecell{90.60 / 92.60 \\ 97.40 / 97.60} 
    & \makecell{96.56 / 96.22 \\ 96.33 / 96.33} 
    & \makecell{53.85 / 74.58 \\ 91.30 / 90.64} 
    & \makecell{76.17 / 80.92 \\ 82.72 / 81.98} 
    & \makecell{78.80 / 84.60 \\ 90.00 / 93.20} 
    & \makecell{91.27 / 89.64 \\ 90.59 / 87.35} 
    & \makecell{95.13 / 93.37 \\ 93.61 / 95.26} 
    & \makecell{84.84 / 83.39 \\ 87.73 / 90.61} 
    & \makecell{91.81 / 88.63 \\ 91.72 / 91.52} 
    & \makecell{85.33 / 87.53 \\ 90.81 / 90.81} & - \\[1ex]
    
    \midrule
    
    \multicolumn{14}{@{}l}{\textit{Llama2-13B / Mistral-7B-v0.2}} \\
    \multicolumn{14}{@{}l}{\textit{Qwen3-14B / DeepSeek-R1-32B}} \\[1ex]
    Pretrain             
    & \makecell{66.61 / 80.46 \\ 83.78 / 88.88} 
    & \makecell{65.00 / 77.76 \\ 84.85 / 72.20} 
    & \makecell{68.20 / 93.80 \\ 96.40 / 93.60} 
    & \makecell{71.33 / 86.24 \\ 88.42 / 91.51} 
    & \makecell{57.19 / 76.92 \\ 88.96 / 92.31} 
    & \makecell{47.42 / 68.71 \\ 80.51 / 82.15} 
    & \makecell{55.60 / 76.80 \\ 85.60 / 89.00} 
    & \makecell{42.03 / 48.43 \\ 81.03 / 80.92} 
    & \makecell{51.47 / 83.95 \\ 82.67 / 54.15} 
    & \makecell{47.65 / 78.70 \\ 83.39 / 64.98} 
    & \makecell{42.65 / 77.64 \\ 77.86 / 62.21} 
    & \makecell{55.92 / 77.22 \\ 84.86 / 79.26} & - \\[2ex]
    LoRA-Tune            
    & \makecell{93.80 / 93.91 \\ 88.67 / 91.93} 
    & \makecell{83.70 / 85.04 \\ 81.50 / 70.09} 
    & \makecell{89.60 / 95.60 \\ 94.00 / 89.40} 
    & \makecell{95.18 / 95.76 \\ 95.41 / 94.72}
    & \makecell{65.89 / 77.59 \\ 91.30 / 92.31} 
    & \makecell{79.12 / 81.16 \\ 81.57 / 84.68} 
    & \makecell{80.40 / 85.60 \\ 90.20 / 94.20} 
    & \makecell{90.91 / 89.27 \\ 90.51 / 86.01} 
    & \makecell{89.84 / 91.93 \\ 92.88 / 91.05} 
    & \makecell{85.56 / 85.20 \\ 89.17 / 86.28} 
    & \makecell{89.65 / 87.16 \\ 90.97 / 90.89} 
    & \makecell{85.79 / 88.02 \\ 89.65 / 88.32} & - \\[2ex]
    Full Fine-tune       
    & \makecell{93.86 / 94.97 \\ 93.46 / 94.45} 
    & \makecell{87.44 / 87.06 \\ 87.25 / 88.11} 
    & \makecell{92.80 / 96.80 \\ 98.40 / 98.40} 
    & \makecell{97.13 / 97.25 \\ 97.59 / 97.36}
    & \makecell{74.92 / 81.94 \\ 93.31 / 94.98} 
    & \makecell{79.12 / 83.78 \\ 87.22 / 90.01} 
    & \makecell{80.40 / 87.00 \\ 90.20 / 95.40} 
    & \makecell{91.73 / 90.43 \\ 91.20 / 91.73} 
    & \makecell{95.83 / 94.60 \\ 94.62 / 96.41} 
    & \makecell{89.17 / 89.17 \\ 91.70 / 94.22} 
    & \makecell{92.01 / 91.15 \\ 93.21 / 93.78} 
    & \makecell{88.58 / 90.38 \\ 92.56 / 94.08} & - \\[1ex]
    
    \midrule
    
    \multicolumn{14}{@{}l}{\textit{Proxy Model Black-Box Tuning Methods}} \\[1ex]
    Proxy-Tune    
    & \makecell{94.12 / 82.38 \\ 82.07 / 89.24} 
    & \makecell{84.08 / 79.00 \\ 69.32 / 80.35} 
    & \makecell{89.40 / 94.20 \\ 86.20 / 79.80} 
    & \makecell{96.79 / 90.48 \\ 90.60 / 90.60}
    & \makecell{57.53 / 77.26 \\ 88.96 / 90.97} 
    & \makecell{75.35 / 70.52 \\ 77.56 / 81.57} 
    & \makecell{78.60 / 77.60 \\ 86.20 / 91.60} 
    & \makecell{90.79 / 74.00 \\ 88.16 / 90.39} 
    & \makecell{95.00 / 90.65 \\ 88.19 / 89.93} 
    & \makecell{82.67 / 80.14 \\ 82.31 / 88.45} 
    & \makecell{89.78 / 84.12 \\ 84.75 / 89.11} 
    & \makecell{84.92 / 81.85 \\ 84.03 / 87.46} & 0\% \\[2ex]
    CPT   
    & \makecell{\textbf{95.45} / \textbf{93.83} \\ \textbf{93.54} / \textbf{93.51}} 
    & \makecell{\textbf{85.91} / \textbf{81.59} \\ 86.29 / 86.67} 
    & \makecell{90.60 / 95.00 \\ 98.20 / 98.40} 
    & \makecell{97.02 / 96.22 \\ 95.87 / 96.10}
    & \makecell{59.87 / 77.59 \\ \textbf{93.31} / \textbf{93.65}} 
    & \makecell{77.89 / 75.18 \\ 85.01 / 87.80}
    & \makecell{\textbf{79.80} / \textbf{83.20} \\ 90.40 / \textbf{93.80}}
    & \makecell{91.01 / 89.15 \\ 90.06 / \textbf{91.41}}
    & \makecell{\textbf{95.28} / 93.26 \\ \textbf{93.78} / \textbf{95.42}}
    & \makecell{85.92 / 80.14 \\ 90.61 / \textbf{90.97}} 
    & \makecell{\textbf{91.81} / 89.07 \\ \textbf{92.28} / \textbf{92.56}} 
    & \makecell{86.41 / 86.75 \\ 91.76 / \textbf{92.75}} & 100\% \\[2ex]
    GP-random (ours)       
    & \makecell{95.30 / 92.20 \\ 92.18 / 92.38} 
    & \makecell{85.43 / 81.21 \\ 85.71 / 86.86} 
    & \makecell{90.60 / 94.80 \\ 98.60 / 98.40} 
    & \makecell{97.02 / 96.10 \\ 95.18 / 95.76}
    & \makecell{58.19 / 78.26 \\ 91.97 / 91.64} 
    & \makecell{78.87 / 73.96 \\ 85.01 / 86.16} 
    & \makecell{79.60 / 82.20 \\ 89.20 / 93.00} 
    & \makecell{90.99 / 89.79 \\ 91.01 / 90.04} 
    & \makecell{94.89 / 92.18 \\ 92.99 / 94.64} 
    & \makecell{86.64 / \textbf{81.95} \\ 90.97 / 90.25} 
    & \makecell{90.46 / 88.60 \\ 91.25 / 91.40} 
    & \makecell{86.18 / 86.48 \\ 91.28 / 91.87} 
    & 6.94\% \\[2ex]
    GP-filter (ours)       
    & \makecell{95.22 / 93.11 \\ 93.22 / 93.08}
    & \makecell{85.81 / 81.02 \\ \textbf{86.77} / \textbf{87.54}} 
    & \makecell{\textbf{92.00} / \textbf{95.40} \\ \textbf{98.80} / \textbf{98.80}} 
    & \makecell{\textbf{97.13} / \textbf{96.44} \\ \textbf{96.10} / \textbf{96.67}} 
    & \makecell{\textbf{61.20} / \textbf{78.26} \\ 92.31 / 91.64}
    & \makecell{\textbf{78.95} / \textbf{75.35} \\ \textbf{86.49} / \textbf{88.62}} 
    & \makecell{79.20 / 82.60 \\ \textbf{90.80} / 93.40}
    & \makecell{\textbf{91.42} / \textbf{90.55} \\ \textbf{91.05} / 90.89} 
    & \makecell{94.98 / \textbf{94.38} \\ 93.57 / 95.22}
    & \makecell{\textbf{87.73} / 81.59 \\ \textbf{92.06} / 89.89} 
    & \makecell{91.71 / \textbf{90.07} \\ 91.93 / 92.09} 
    & \makecell{\textbf{86.85} / \textbf{87.16} \\ \textbf{92.10} / 92.53} 
    & \makecell{\textbf{1.38\%} / \textbf{1.45\%} \\ \textbf{1.58\%} / \textbf{1.51\%}} \\[1ex]
    \bottomrule
  \end{tabular}
  }
  \caption{Experimental results comparing our GP tuning with other approaches, including white-box LoRA and black-box proxy tuning methods, across 11 datasets. Our techniques are denoted by GP-random and GP-filter. 
  We separately use \textit{Llama2-7B, Mistral-7B-Instruct-v0.1, Qwen3-8B, DeepSeek-R1-Distill-Qwen-14B} as a small white-box proxy model, and \textit{Llama2-13B, Mistral-7B-Instruct-v0.2, Qwen3-14B, DeepSeek-R1-Distill-Qwen-32B} as the black-box foundation model. 
  “Pretrain" refers to zero-shot inference using official pretrained parameters, “LoRA-Tune" denotes fine-tuning via LoRA \cite{hu2022lora}, and “Full Fine-tune" refers to directly fine-tuning all model parameters.
  The methods Proxy-Tune, CPT, GP-random, and GP-filter are grouped as \textit{Proxy Model Black-Box Tuning Methods}. All datasets are evaluated by Accuracy (higher is better).}
  \label{tab:experiment-results}
\end{table*}

\begin{table*}[t]
  \centering
  \renewcommand{\arraystretch}{1.5}
  \resizebox{\linewidth}{!}{
  \begin{tabular}{lcccccccccccc}
    \toprule
    \multirow{2}{*}{Model \& Method} & \multicolumn{12}{c}{\textbf{API Call Efficiency (\%) $\downarrow$}} \\
    \cmidrule(lr){2-13}
        & AG-News & CoLA  & CoPA  & SST-2 & ARC-C & Cs-QA & OB-QA & MNLI  & QNLI  & RTE   & QQP   & Avg.  \\
    \midrule
    CPT \cite{he2024cpt}   & 100.00 & 100.00 & 100.00 & 100.00 & 100.00 & 100.00 & 100.00 & 100.00 & 100.00 & 100.00 & 100.00 & 100.00 \\
    GP-random (ours)       & 3.33   & 11.69  & 10.00  & 5.94   & 9.83   & 10.27  & 8.07   & 1.02   & 4.77   & 10.04  & 1.37   & 6.94 \\[0.6ex]
    GP-filter (ours)       
    & \makecell{1.08 / 0.87 \\ 1.21 / 1.25}   
    & \makecell{0.09 / 0.18 \\ 0.74 / 0.20}   
    & \makecell{0.60 / 1.60 \\ 1.20 / 1.90}  
    & \makecell{1.07 / 1.02 \\ 1.38 / 1.27}
    & \makecell{3.31 / 2.14 \\ 2.95 / 2.86}
    & \makecell{1.95 / 3.07 \\ 2.77 / 1.98} 
    & \makecell{0.87 / 1.43 \\ 2.08 / 1.39}
    & \makecell{0.70 / 0.78 \\ 0.83 / 0.81}   
    & \makecell{1.92 / 1.13 \\ 1.79 / 1.83}
    & \makecell{2.17 / 2.21 \\ 1.08 / 2.13}
    & \makecell{1.39 / 1.48 \\ 1.37 / 1.04}
    & \makecell{\textbf{1.38} / \textbf{1.45} \\ \textbf{1.58} / \textbf{1.51}} \\
    \bottomrule
  \end{tabular}
  }
  \caption{Large model API call efficiency. This table compares the percentage of large model API calls used by our methods versus CPT \cite{he2024cpt}. For CPT, an API call is made for every training instance (i.e., 100\% usage). Our GP-filter method requires only an average of \textbf{1.38\%} (\textit{Llama2} family models) of these calls.}
  \label{tab:api-call}
\end{table*}

\paragraph{Models Selection.} 
To balance efficiency and reliability, we select models from \textbf{Llama2}~\cite{touvron2023llama}, \textbf{Mistral-7B}~\cite{jiang2023mistral7b}, \textbf{Qwen3}~\cite{yang2025qwen3technicalreport} and \textbf{DeepSeek-R1-Distill}~\cite{deepseekai2025deepseekr1incentivizingreasoningcapability} families. For each series, we use a small model as the proxy and a larger one as the target. While the selected large models are technically white-box, we treat them as black-box during validation to simulate realistic constraints. We also directly fine-tune them to obtain an oracle upper bound, which serves as a reference for evaluating our method's effectiveness. For evaluations involving truly inaccessible black-box models, please refer to Appendix, where we conduct experiments under genuine black-box conditions.

\paragraph{Datasets Selection.} 
We conducted experiments across diverse NLP datasets to showcase the versatility of our method. Our approach was evaluated on three major tasks: \textbf{(a).} Text Classification: We use AG-News \cite{zhang2015character}, CoLA (Corpus of Linguistic Acceptability) \cite{warstadt2019neural}, SST-2 (Stanford Sentiment Treebank) \cite{socher2013recursive}, and QQP (Quora Question Pairs) \cite{qqp}. \textbf{(b).} Question Answering: We include ARC-C (AI2 Reasoning Challenge - Challenge Set) \cite{clark2018think}, Cs-QA (CommonsenseQA) \cite{talmor2018commonsenseqa}, and OB-QA (OpenBookQA) \cite{mihaylov2018can}. \textbf{(c).} Natural Language Inference: We consider MNLI (Multi-Genre Natural Language Inference) \cite{williams2017broad}, QNLI (Question Natural Language Inference) \cite{rajpurkar2018know}, RTE (Recognizing Textual Entailment) \cite{dagan2005pascal}, and CoPA (Choice of Plausible Alternatives) \cite{roemmele2011choice}. These datasets are widely used and well-established benchmarks, covering diverse linguistic phenomena and evaluation challenges.

\paragraph{Baselines.}
We compare our proposed GP-based proxy tuning method for black-box LLMs against several representative baselines to demonstrate its effectiveness: \textbf{(a).} Zero-shot Inference: We evaluate the pretrained LLMs without any tuning, by directly applying them to the test sets. This provides a fundamental reference point for all tuning methods. \textbf{(b).} Direct Fine-tuning: We apply both LoRA \cite{hu2022lora} and full-precise fine-tuning on the LLMs. Although these methods require full access to model parameters, which is impractical in black-box scenarios, they provide upper-bound references to demonstrate the performance of our approach. \textbf{(c).} Proxy-Tuning and CPT: We compare our method with leading black-box tuning approaches, including Proxy-Tuning \cite{liu2024tuning} and CPT \cite{he2024cpt}, demonstrating that our approach achieves superior performance while requiring substantially fewer API calls.

\paragraph{Empirical Data Selection.}
\label{recommendation}
As shown in Table~\ref{tab:api-call}, we empirically determined the data proportions for both random and filter selection strategies. For the random strategy, we iteratively adjusted the sampling ratio until the resulting accuracy matched that of CPT. For the filter strategy, we tuned the input and output thresholds, and repeatedly applied Algorithm~\ref{algo} to construct LogitMap Pairs, aiming to strike a balance between efficiency and performance.

Based on extensive empirical evaluation, we make the following recommendations:

\begin{itemize}
    \item \textbf{Random-based:} For datasets fewer than 100K samples, 5\% random sampling performs well. For larger datasets, sampling 5K examples is sufficient.
    \item \textbf{Filter-based:} For most datasets, selecting around 1\% of the data via our filtering algorithm is sufficient to achieve strong performance. However, for extremely large datasets such as MNLI and QQP, which contain nearly 400K samples, fitting a GP model on a proportional subset becomes computationally infeasible and may lead to numerical instability (e.g., NaN predictions). To address this, we recommend sampling approximately 2K examples, which balances efficiency and accuracy.
\end{itemize}

Compared to tuning the small proxy model, the phase of constructing LogitMap Pairs and Training Gaussian Process model is significantly more efficient in both time and memory. \textbf{Detailed results} are provided in Appendix. 

\subsection{Main Results}
The main experimental results are presented in Table~\ref{tab:experiment-results}. 
Our proposed GP-based tuning methods, particularly GP-filter, exhibit consistently strong performance across all evaluated models and datasets. Focusing first on the \textit{Llama2} family, GP-filter improves the average accuracy from \textbf{55.92\%} (pretrained) to \textbf{86.85\%} (GP-filter tuned), even outperforming LoRA-Tune (85.79\%) and approaching the performance of full fine-tuning (88.58\%) on \textit{Llama2-13B}. Compared to other proxy-based approaches, GP-filter achieves a higher average accuracy than CPT (86.41\%) while using only \textbf{1.38\%} of its API calls—to the best of our knowledge, this represents the state-of-the-art in both performance gain and API efficiency—demonstrating both effectiveness and remarkable cost-efficiency. It also outperforms offline Proxy-Tune by an average margin of 1.93 percentage points.

To validate generalizability, we conduct extensive experiments across other model families including \textit{Mistral-7B}, \textit{Qwen3}, and \textit{DeepSeek-R1-Distill}. In all cases, GP-filter yields consistent improvements over the pretrained models, achieving average accuracy gains of \textbf{+9.94}, \textbf{+7.24}, and \textbf{+13.27} percentage points, respectively. Notably, in each setting, GP-filter uses few API calls (below 2\%), highlighting its extreme cost-efficiency alongside strong performance.

We also apply GP-filter in a simulated \textbf{real-world black-box LLM} setting (\textit{Qwen-Plus} model from Tongyi Qianwen) to further assess its practical applicability. 
As detailed in Appendix \ref{sec:realblackbox}, the method remains highly effective even under multiple realistic constraints, confirming its robustness in truly black-box environments.

\subsection{Ablation Study: Different Usage of API Calls}
\begin{figure}[h!]
  \centering
  \begin{subfigure}[b]{\columnwidth}
    \centering
    \includegraphics[width=\columnwidth]{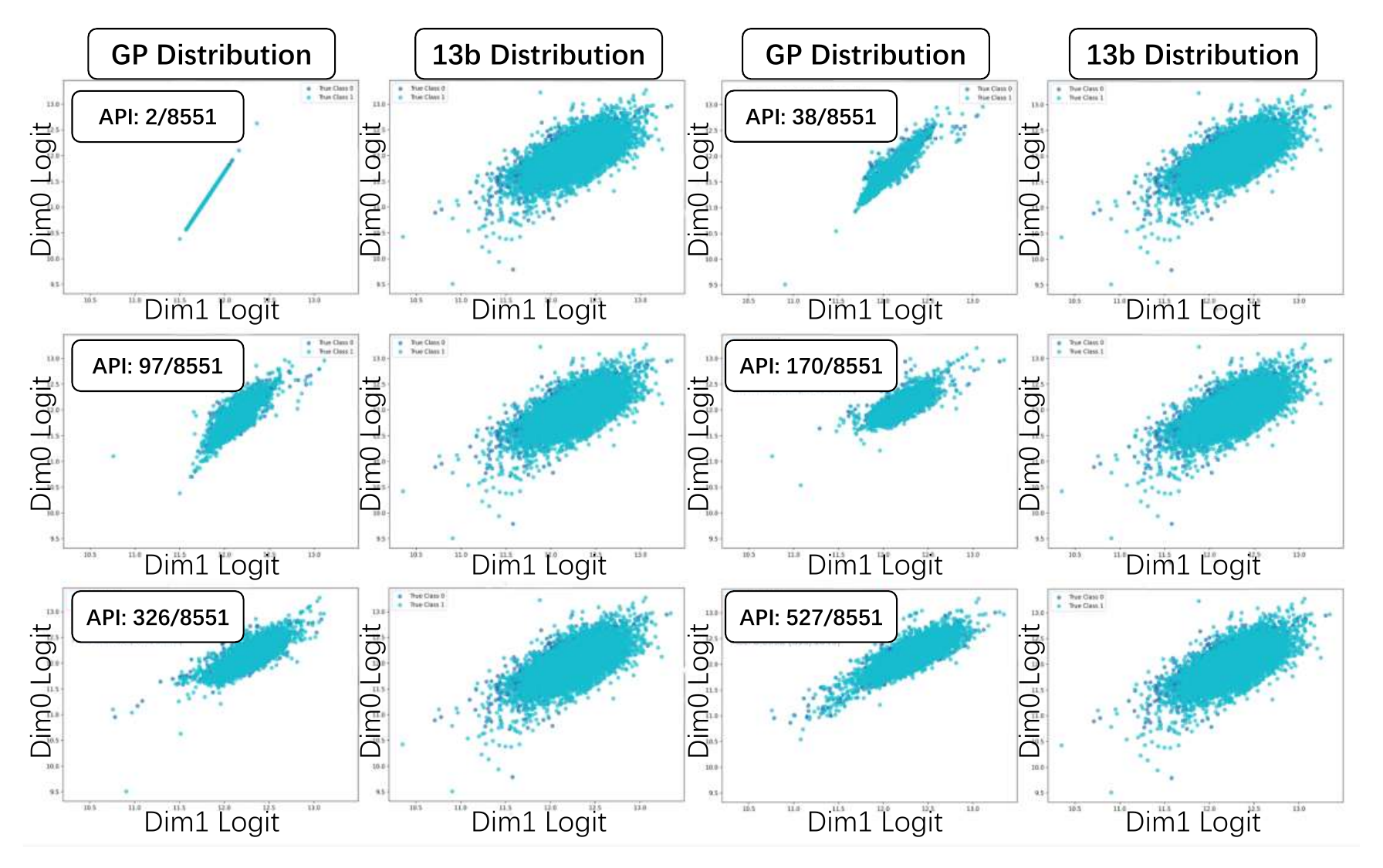}
    \caption{CoLA: across six API call budgets}
    \label{fig:cola}
  \end{subfigure}
  \hspace{0.005\textwidth}
  \begin{subfigure}[b]{\columnwidth}
    \centering
    \includegraphics[width=\columnwidth]{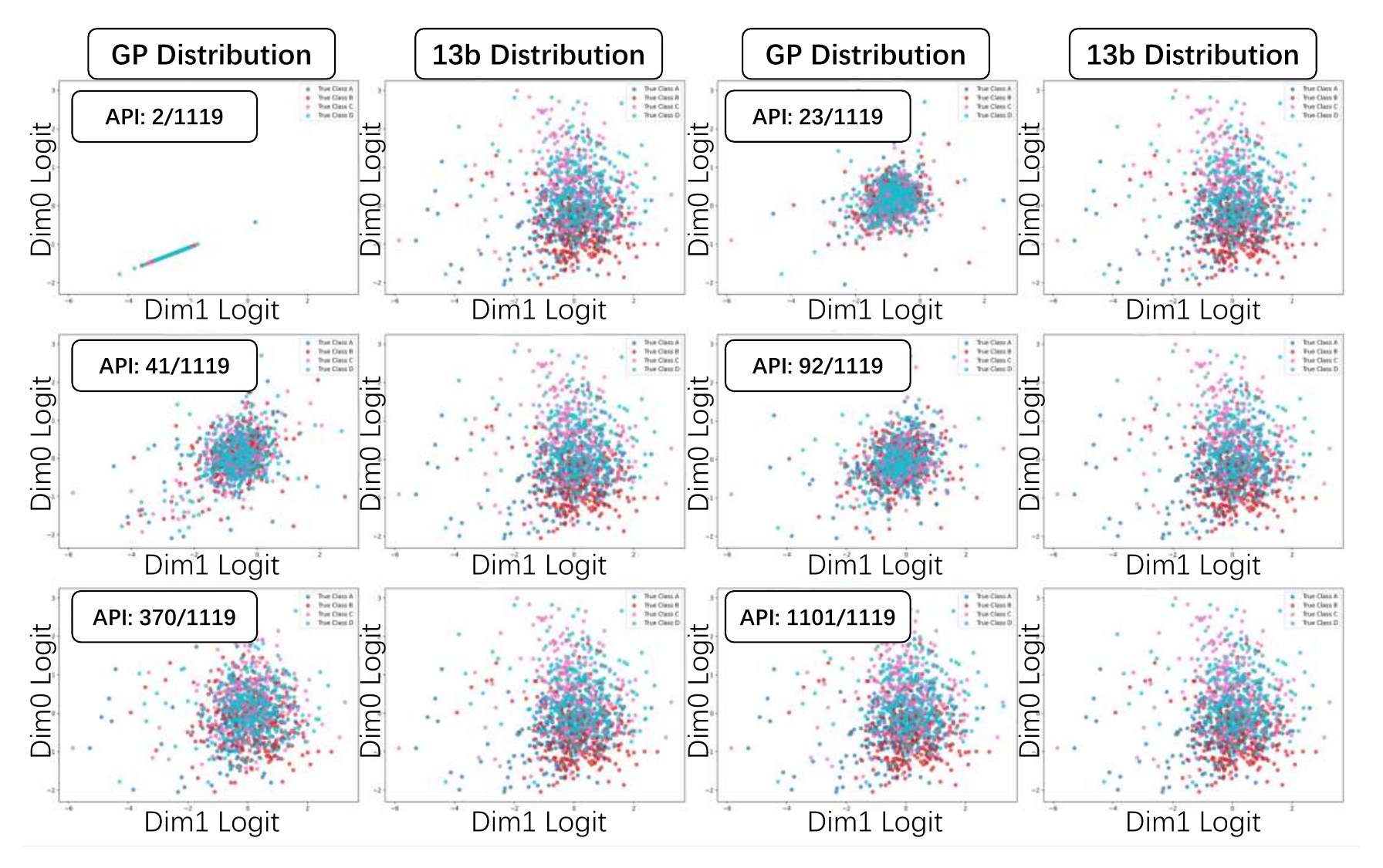}
    \caption{ARCC: across six API call budgets}
    \label{fig:arcc}
  \end{subfigure}
  \caption{Logits distributions produced by the GP models (each left subfigures) and the target black-box model (each right subfigures) across two datasets. The figure consists of four columns, grouped as two pairs: the first and third columns show the output logits distributions of six GP models trained under different API budgets; the second and fourth columns display the output logits of the same \textit{Llama2-13b} pretrained model under the same inputs, repeated for clearer comparison. The 2 and 4 colors in (a) and (b) represent the 2 and 4 classes in their respective datasets.}
  \label{fig:logits_dist}
\end{figure}

We trained GP models on 6 datasets—CoPA, ARCC, CoLA, RTE, OBQA, and MRPC—using different API calls to generate variant GP surrogates, which were then used to guide the small proxy via the GP-filter method. Figure~\ref{fig:logits_dist} compares the output logits distributions of GP model and the target foundation model for CoLA and ARCC. Results for other datasets are provided in Appendix~\ref{sec:distribution}. 
In each subfigure, the left panel shows the GP’s output distribution, while the right panel shows that of the \textit{Llama2-13B} pretrained model.

Under extreme data scarcity, the GP logits distribution becomes highly compressed, as observed in Figure~\ref{fig:cola} (the top left subfigure). With \textbf{only} 2 API calls (out of 8,551), the distribution nearly degenerates into a one-dimensional form, effectively reducing the GP model to a linear function. Surprisingly, the performance of the GP-filter method remains robust. In contrast, Figure~\ref{fig:arcc} (the bottom right subfigure) shows the opposite extreme, where nearly all available data (1,101 out of 1,119) is used. In this case, GP logits closely match those of the \textit{Llama2-13B} pretrained model, and the GP-filter method achieves performance comparable to CPT.

These results highlight that GP-filter can perform well even with \textbf{extremely limited API calls}. Rather than replicating the large model's logits, the GP appears to approximate its underlying knowledge structure, with added noise that may serve as implicit regularization, enhancing the proxy’s robustness. Even in extreme minimal-data regimes (e.g., 2 samples in CoLA), the GP model provides a useful corrective signal, capturing high-level structural patterns such as distributional tendencies and relative logit relationships.

Additional ablation studies provide further insights into the strengths of our approach. As shown in appendix~\ref{ablation-small}, 
under extreme data scarcity, directly fine-tuning the \textit{Llama2-7B} model yields low effectiveness. In contrast, our GP-filter method maintains high performance, achieving 6.31 percentage points higher accuracy. Furthermore, appendix~\ref{ablation-hard} 
shows that on more challenging datasets, our method significantly outperforms offline approaches such as Proxy-Tune. This improvement stems from the fact that Proxy-Tune only leverages the foundation model at the reference stage; however, in difficult tasks, both the fine-tuned proxy model and the pretrained large model may perform suboptimally. In contrast, our approach incorporates the foundation model’s knowledge throughout the training process, leading to more effective proxy fine-tuning and consistently better results.

\section{Conclusion}
\label{conclusion}
In this paper, we introduce a Gaussian Process (GP) based proxy tuning approach that directly addresses the central challenge in black-box LLM adaptation: aligning models effectively while minimizing expensive API interactions. Our approach trains a GP surrogate on a small, curated dataset to approximate the target model’s behavior and uses its predictive uncertainty to guide selective querying during proxy fine-tuning. 
Extensive experiments demonstrate that our method matches or outperforms competitive online black-box tuning techniques while using substantially fewer queries, and consistently surpasses existing offline strategies. These results not only highlight the practicality and cost-efficiency of GP-based proxy tuning, but also confirm its effectiveness in realistic black-box adaptation scenarios.

\section{Acknowledgments}
This work was supported by the National Natural Science Foundation of China Youth Student Basic Research Program (Grant No. 625B1002). It was also supported by High-Quality Development Project of Shanghai Municipal Commission of Economy and Informatization (Grant No. 2024-GZL-RGZN-02010) and AI for Science Foundation of Fudan University (FudanX24AI028).

\bibliography{aaai2026}

\clearpage

\appendix
\renewcommand{\thesection}{\Alph{section}.\arabic{section}}
\setcounter{section}{0}
\setcounter{table}{0}
\setcounter{figure}{0}
\renewcommand{\thetable}{S\arabic{table}}
\renewcommand{\thefigure}{S\arabic{figure}}

\makeatletter

\renewcommand{\thesection}{\Alph{section}}
\setcounter{secnumdepth}{2}
\makeatother


\textbf{This appendix can be divided into six parts:}
\begin{enumerate}
    \item Section \ref{sec:implementation} provides details on the experimental setup, including model selection, prompt engineering under black-box constraints, and evaluation configurations.

    \item Section \ref{sec:hyper} presents the heuristic selection strategies and sensitivity analyses for key hyperparameters, including: (1) the input / output thresholds (\(\tau_{\mathrm{in}}, \tau_{\mathrm{out}}\)) in \ref{distance} , (2) the re-querying gate threshold (\(\theta\)) in \ref{theta} , and (3) the logit difference weight (\(\alpha\)) in \ref{alpha} .
    
    \item Section \ref{sec:distribution} presents supplementary visualizations comparing the GP surrogate's output logits distributions with those of the target foundation model.

    \item Section \ref{sec:appendix_ablations} presents more ablation results, showcasing the strength of GP-filter method through: (1) strong performance under extreme data scarcity in \ref{ablation-small}, (2) significant advantages on challenging datasets in \ref{ablation-hard}, and (3) robustness across varying API call budgets in \ref{ablation-apiusage}.

    \item Section~\ref{sec:gp-ex} provides time and memory analysis for LogitMap pair construction and GP training, demonstrating that our GP training phase is highly efficient and incurs minimal computational overhead.
    
    \item Section~\ref{sec:realblackbox} conducts real-world black-box tuning experiments with Qwen-Plus API, showing the practical effectiveness of the our method under realistic constraints.
\end{enumerate}

\section{Implementation Details}
\label{sec:implementation}

\paragraph{Prompt Construction} 
As described in the main text, we separately employ \textit{Llama2-7B, Mistral-7B-Instruct-v0.1, Qwen3-8B, DeepSeek-R1-Distill-Qwen-14B} as a small proxy model and \textit{Llama2-13B, Mistral-7B-Instruct-v0.2, Qwen3-14B, DeepSeek-R1-Distill-Qwen-32B} as black-box foundation model. Although both models are inherently generative, many of our target tasks are classification-oriented. In strict black-box settings, we do not have access to model internals—such as modifying the architecture to append classification heads or accessing final-layer parameters. To accommodate this constraint, we design task-specific prompts that elicit the model to generate outputs consistent with the task’s classification labels.

For instance, in multiple-choice tasks, we craft prompts that guide the model to output a single option token (e.g., “A", “B", “C", etc.). For binary classification tasks (e.g., true / false), the model is prompted to respond with “Yes" or “No". This approach allows us to interpret predictions by inspecting the logits of specific output tokens, without modifying the model architecture, thus strictly adhering to the black-box setting.

Each prompt is composed of several components derived from the target dataset:
\begin{itemize}
    \item \{sentence1\}, \{sentence2\} or \{text\}: The main context.
    \item \{question\}: The main question context.
    \item \{options\}: The reasonable choices for the task (From original dataset or designed by us).
    \item \{answer\}: The correct answer, which is included during training but omitted during testing.
\end{itemize}
Our prompt details for 11 datasets are given in Table \ref{tab:prompt_info}. 

\begin{table}[t]
    \centering
    \resizebox{\linewidth}{!}{
    \begin{tabular}{p{2cm} p{1.5cm} p{1.5cm} p{8cm}}
        \toprule
        Dataset & Train Size & Test Size & Prompt Construction Details \\
        \midrule
        AG-News & 120,000 & 7,600 &
        \multicolumn{1}{p{8cm}}{\raggedright 
        This is a news article: \{text\} \\ 
        Question: \{question\} Options: \{options\} \\
        Answer: \{answer\}} \\
        
        \midrule
        CoLA & 8,551 & 1,043 &
        \multicolumn{1}{p{8cm}}{\raggedright 
        Sentence: \{text\} \\
        Determine whether the given sentence is grammatically correct according to standard English grammar rules. Respond with a single word: 'Yes' if the sentence is grammatically correct, or 'No' if it is not. \\
        Answer: \{answer\}} \\
        
        \midrule
        CoPA & 1,000 & 500 &
        \multicolumn{1}{p{8cm}}{\raggedright 
        Premise: \{text\} \\
        Question: What was the cause (or effect) ? \\
        Options: \{options\} \\
        Answer: \{answer\}} \\

        \midrule
        SST-2 & 67,349 & 872 &
        \multicolumn{1}{p{8cm}}{\raggedright 
        Sentence: \{text\} \\
        Question: What is the sentiment of the sentence? \\
        Options: \{options\} \\ 
        Answer: \{answer\}} \\
        
        \midrule
        ARC-C & 1,119 & 299 &
        \multicolumn{1}{p{8cm}}{\raggedright 
        Question: \{question\} Options: \{options\} \\ 
        The correct answer is: \{answer\}} \\

        \midrule
        Cs-QA & 9,741 & 1,221 &
        \multicolumn{1}{p{8cm}}{\raggedright 
        Question: \{question\} Options: \{options\} \\ 
        The correct answer is: \{answer\}} \\

        \midrule
        OB-QA & 4,957 & 500 &
        \multicolumn{1}{p{8cm}}{\raggedright 
        Background Fact: \{text\} \\
        Question: \{question\} Options: \{options\} \\ 
        The correct answer is: \{answer\}} \\
        
        \midrule
        MNLI & 392,702 & 9,815 &
        \multicolumn{1}{p{8cm}}{\raggedright 
        Premise: \{sentence1\} Hypothesis: \{sentence2\} \\
        Question: What is the relationship between the Premise and the Hypothesis? Options: \{options\} \\ 
        Please choose one option (A, B, or C) as your answer. Answer: \{answer\}} \\

        \midrule
        QNLI & 104,743 & 5,463 &
        \multicolumn{1}{p{8cm}}{\raggedright 
        Question: \{question\} \\
        Sentence: \{text\} \\
        Does the sentence entail the answer to the question? \\ 
        Reply with 'Yes' or 'No'. Answer: \{answer\}} \\

        \midrule
        RTE & 2,490 & 277 &
        \multicolumn{1}{p{8cm}}{\raggedright 
        Premise: \{sentence1\} Hypothesis: \{sentence2\} \\
        Does the premise entail the hypothesis? \\ 
        Reply with 'Yes' or 'No'. Answer: \{answer\}} \\

        \midrule
        QQP & 363,846 & 40,430 &
        \multicolumn{1}{p{8cm}}{\raggedright 
        Question1: \{sentence1\}  Question2: \{sentence2\} \\
        Question: Are these two questions semantically equivalent? (Yes or No) \\
        Answer: \{answer\}} \\
        \bottomrule
    \end{tabular}
    }
    \caption{Prompt Construction for Each Dataset.}
    \label{tab:prompt_info}
\end{table}

\paragraph{Experimental Methods and Configurations}
All experiments are conducted using the PyTorch framework, with comprehensive evaluations and comparisons of multiple tuning approaches: (1) LoRA-tuning and full-precision direct tuning on both small proxy and large black-box model. (2) Proxy-based tuning methods for black-box settings, including Proxy-Tuning (PT)~\cite{liu2024tuning}, Consistent Proxy Tuning (CPT)~\cite{he2024cpt}, and our proposed Gaussian Process-based proxy tuning method, with both random sampling and rule-based filtering strategies.

To ensure a fair and reproducible comparison, we use consistent training settings across all proxy-based tuning methods. All experiments are conducted on a single machine equipped with 6 NVIDIA GeForce RTX 4090 GPUs. We utilize mixed-precision training (fp16) for efficiency. The key hyperparameters are summarized in Table~\ref{tab:exp_config}.

\begin{table}[h!]
\centering
\resizebox{\linewidth}{!}{
\begin{tabular}{ll}
\toprule
\textbf{Setting} & \textbf{Value} \\
\midrule
Number of GPUs & 6 (NVIDIA GeForce RTX 4090) \\
Training epochs & 2 \\
Batch size per device & 4 \\
Gradient accumulation steps & 8 \\
Optimizer & paged\_adamw\_32bit \\
Learning rate & $2 \times 10^{-4}$ \\
Learning rate scheduler & Linear \\
Precision & Mixed precision (fp16) \\
torch.dtype & torch.float16 \\
\bottomrule
\end{tabular}
}
\caption{Experimental settings details.}
\label{tab:exp_config}
\end{table}


\section{Hyperparameter Selection}
\label{sec:hyper}

\subsection{Input and Output Thresholds \(\tau_{\mathrm{in}}, \tau_{\mathrm{out}}\)}
\label{distance}
The thresholds \(\tau_{\mathrm{in}}\) and \(\tau_{\mathrm{out}}\) are hyperparameters for our GP filtering method (Algorithm~\ref{algo}). \(\tau_{\mathrm{in}}\) is the criterion for the input embedding space, and \(\tau_{\mathrm{out}}\) is the criterion for the output logits. These thresholds jointly govern the filtering stringency for selecting LogitMap Pairs, which form the training dataset for our GP surrogate model.

To determine these thresholds heuristically, we conduct an initial calibration run of Algorithm~\ref{algo} using permissive (i.e., very low) values for \(\tau_{\mathrm{in}}\) and \(\tau_{\mathrm{out}}\). This allows the collection of scalar metric values from all potential input embeddings (forming set \(M_{\mathrm{in}}\)) and their corresponding output logits (forming set \(M_{\mathrm{out}}\)). After sorting these metrics, we set \(\tau_{\mathrm{in}}\) and \(\tau_{\mathrm{out}}\) to the 1st percentile of the sorted values in \(M_{\mathrm{in}}\) and \(M_{\mathrm{out}}\), respectively. This percentile-based heuristic offers a simple yet effective way to adaptively set \(\tau_{\mathrm{in}}\) and \(\tau_{\mathrm{out}}\). In our experiments, this strategy results in only \textbf{\(\approx 0.7\%\)} of API calls being used during the filtering process on average, while still preserving high data diversity. 

\subsection{Gate Threshold $\theta$}
\label{theta}

\begin{figure}[h]
    \centering
    \includegraphics[width=\linewidth]{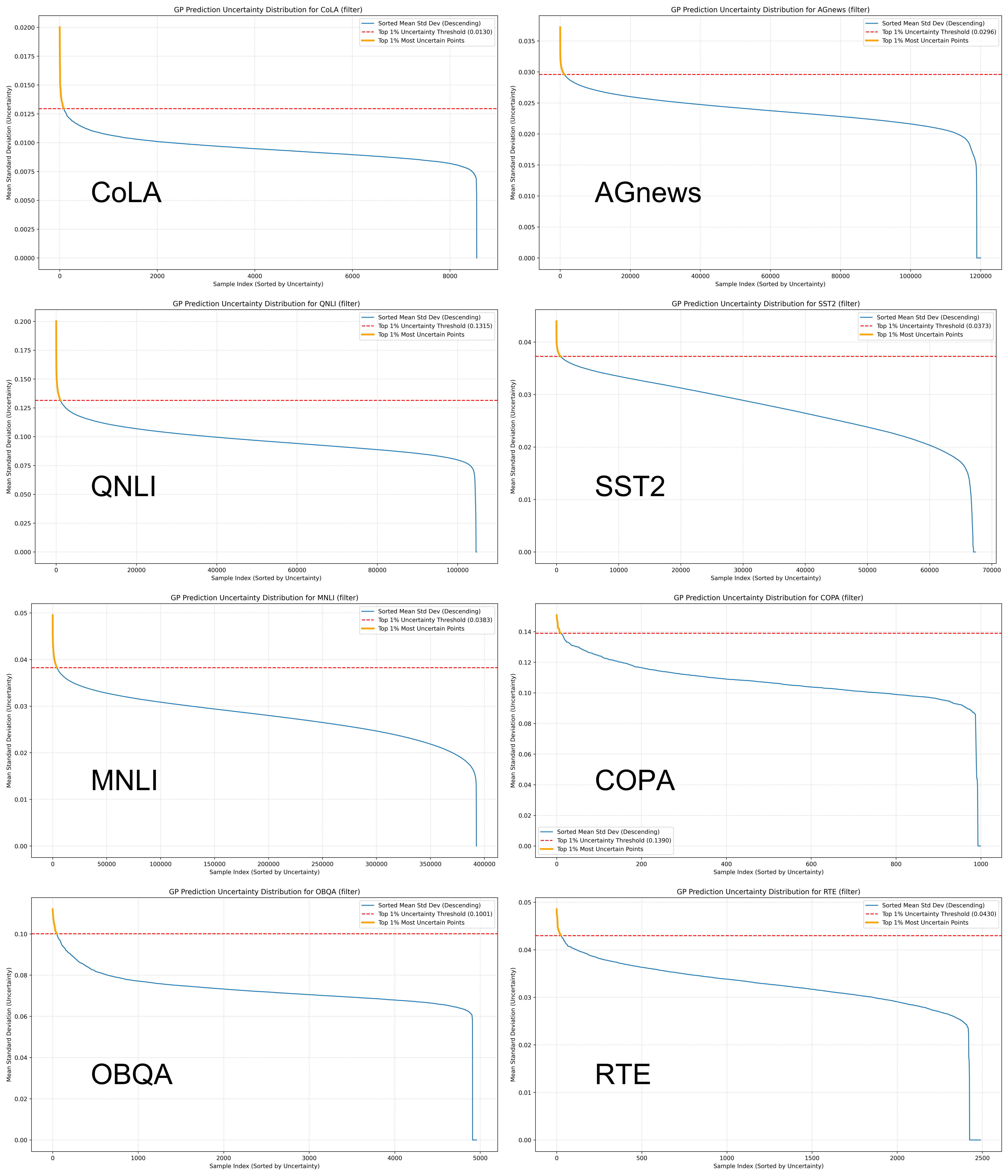}
    \caption{Visualization of sorted Gaussian Process prediction uncertainties ($\tau_{\mathcal{M}_{gp}}(x)$) across 8 benchmark datasets: CoLA, AG-News, QNLI, SST-2, MNLI, COPA, OBQA, and RTE. The orange segment in each plot highlights the top 1\% of queries with the highest uncertainty, prompting the system to query the large model $\mathcal{M}_l$. The red dashed line marks the 1st percentile uncertainty value, serving as the gate threshold $\theta$, which consistently aligns with a sharp decline or “elbow" in the uncertainty distribution.}
    \label{fig:gp_uncertainty_visualization}
\end{figure}

To efficiently manage API calls during proxy model \(\mathcal{M}^+_s\) fine-tuning, we employ an uncertainty-guided re-querying strategy. When GP model predictions are highly uncertain (i.e., exhibit a large standard deviation \(\sigma_{\mathcal{M}_{gp}}(x)\)), a gating threshold \(\theta\) determines whether to consult \(\mathcal{M}_l\). This \(\theta\) is heuristically initialized at the 1st percentile of sorted GP uncertainty values, thereby targeting the top 1\% of the most uncertain training queries for re-evaluation by \(\mathcal{M}_l\). This adaptive mechanism aims to mitigate noise from low-confidence GP predictions, preserving the quality of supervision and enhancing both training stability and task performance.

The selection of the 1st percentile as this initial threshold is empirically validated by the characteristic distribution of GP uncertainties across datasets, as illustrated in Figure~\ref{fig:gp_uncertainty_visualization}. These distributions consistently show a steep decline in uncertainty for a small fraction of examples (typically the top 1\%), followed by a gradual decrease or plateau. Notably, this 1st percentile threshold (demarcated by the orange segment in Figure~\ref{fig:gp_uncertainty_visualization}) often coincides with a discernible elbow or inflection point in the uncertainty curves of many datasets (e.g., CoLA, AG-News, QNLI, SST-2, and MNLI). For other datasets (COPA, OBQA, RTE), this top 1\% region also reliably isolates instances of acutely high, rapidly diminishing uncertainty. These observations underscore that this top 1\% subset generally comprises the least reliable GP predictions, making them primary candidates for re-querying via \(\mathcal{M}_l\).

Building upon the 1st percentile as an empirically grounded default, we further refine API expenditure by optimizing the re-querying budget per dataset. Starting from this 1\% target, we reduce the re-querying volume while monitoring task performance to prevent degradation. The CoLA dataset exemplifies this strategy's efficacy: re-querying only \(\approx 0.07\%\) of its most uncertain examples during fine-tuning—coupled with an initial data construction cost of just \(\approx 0.02\%\)—delivered strong results with a total API overhead of merely \(\approx 0.09\%\). On average, across all datasets, this active re-querying phase during fine-tuning accounts for only \(\approx 0.7\%\) of API calls. This adaptive, cost-aware methodology achieves an effective balance between computational efficiency and high-quality supervision, fostering reliable proxy model training with minimal overhead.

\subsection{Weight of Logit Difference \(\alpha\)}
\label{alpha}
The hyperparameters \(\alpha_{\textit{train}}\) and \(\alpha_{\textit{test}}\) control the influence of the logit difference term, which quantifies the divergence between the logits of the primary model (GP model or large black-box model) and the untrained proxy model. We consistently set \(\alpha_{\textit{train}}\) to 0.8 across all datasets in experiments. For main results reported in Table~\ref{tab:experiment-results}, \(\alpha_{\textit{test}}\) was also set to 0.8 to maintain consistency with the training configuration.

\begin{figure}[h]
    \centering
    \includegraphics[width=\linewidth]{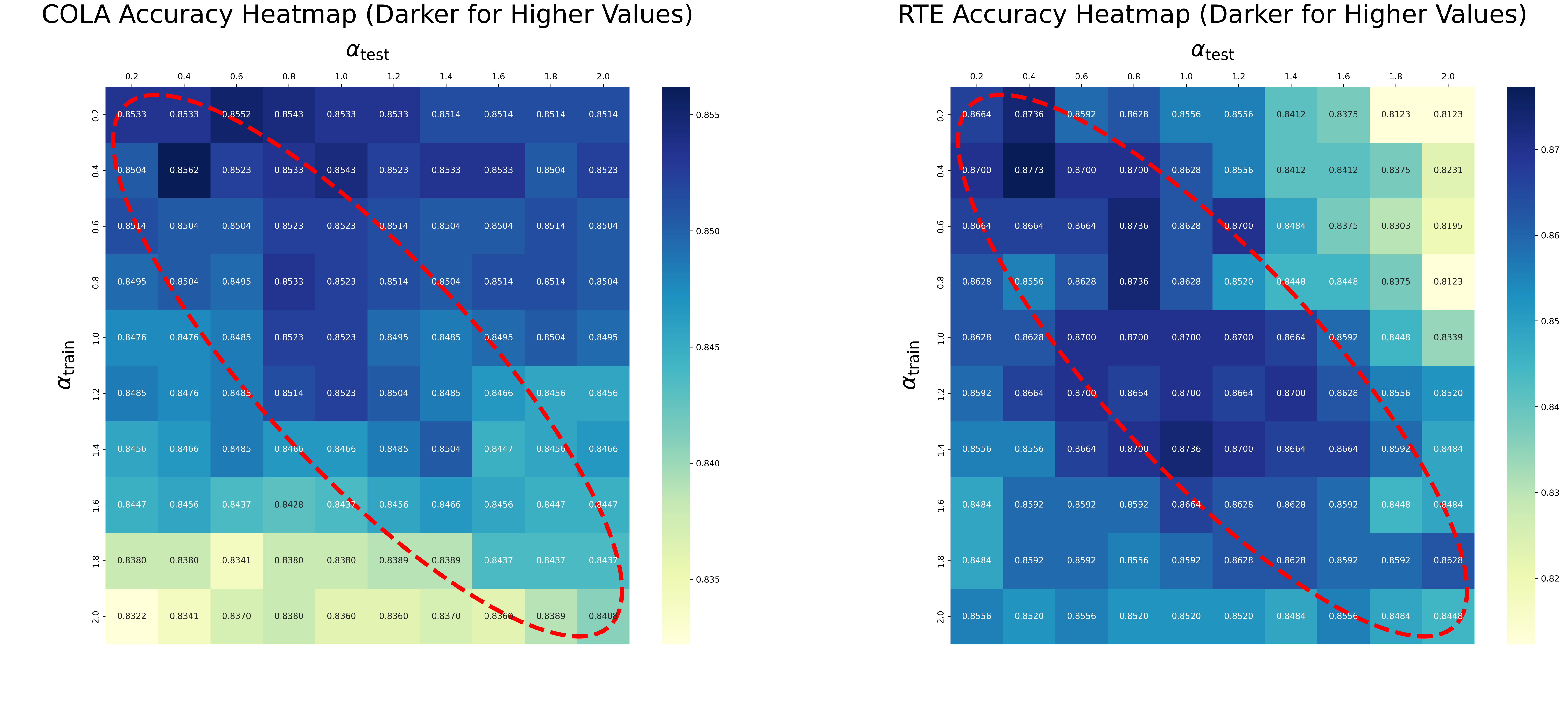}
    \caption{Heatmaps illustrating the performance (i.e., accuracy scores, where darker shades indicate higher values) on the CoLA and RTE for various combinations of \(\alpha_{\textit{train}}\) and \(\alpha_{\textit{test}}\). The red ellipses highlight regions where \(\alpha_{\textit{train}} \approx \alpha_{\textit{test}}\), generally corresponding to optimal performance.}
    \label{fig:alpha_heatmap_cola_rte}
\end{figure}

To assess the sensitivity of our method to different \(\alpha\) values, we conducted additional experiments on the CoLA and RTE using our GP-filter method. The results, visualized as heatmaps in Figure~\ref{fig:alpha_heatmap_cola_rte}, reveal several key trends.

Firstly, the highest performance scores are generally achieved when \(\alpha_{\textit{train}}\) and \(\alpha_{\textit{test}}\) are aligned (i.e., their values are approximately equal). This corresponds to the regions along the main diagonal of the heatmaps, highlighted by the red ellipses. This finding underscores the importance of maintaining consistency in the weighting of the logit difference term between the training and inference phases.

Secondly, while aligning \(\alpha_{\textit{train}}\) and \(\alpha_{\textit{test}}\) is generally beneficial, an important pattern emerges when observing performance along the main diagonal of the heatmaps (Figure~\ref{fig:alpha_heatmap_cola_rte}). Specifically, as both \(\alpha_{\textit{train}}\) and \(\alpha_{\textit{test}}\) concurrently increase to very high values (bottom-right corner), an obvious degradation in performance is observed. This decline appears to stem primarily from two interconnected factors:

\begin{itemize}
    \item \textbf{Over-reliance on the Large Model.} When \(\alpha\) becomes excessively large, the proxy model may become overly influenced by the large model's outputs during fine tuning, leading to an over-reliance on its guidance. This, in turn, can diminish the supervision of the fine tuning data, making it difficult for proxy to capture task-specific nuances and generalize effectively.
    \item \textbf{Propagation of Large Model's Noise.} When the large model’s output logits are inherently noisy or exhibit high uncertainty for data instances, an overly large \(\alpha\) may lead the proxy model to internalize this noise. Moreover, the large model's guidance can substantially deviate from the true output distribution, thereby disrupting the proxy's learning process and ultimately undermining its predictive stability and overall accuracy.
\end{itemize}

These observations indicate that both aligning and properly scaling \(\alpha_{\textit{train}}\) and \(\alpha_{\textit{test}}\) are critical. The proxy model should be able to benefit from the large model’s guidance without compromising its own learning capacity. Based on our empirical analyses (as illustrated in Figure~\ref{fig:alpha_heatmap_cola_rte}), we recommend selecting \(\alpha\) values within the approximate range of [0.6, 1.4]. Across all our main experiments, setting \(\alpha_{\textit{train}} = \alpha_{\textit{test}} = 0.8\) consistently provided a strong balance and robust performance across all evaluated datasets.


\section{Comparison of Output Logits Distributions for Additional Datasets}
\label{sec:distribution}

This appendix provides supplementary visualizations for the ablation study on API call usage presented in the main text.

\begin{figure}[H]
  \centering
  \begin{subfigure}[t]{\linewidth}
    \centering
    \includegraphics[width=0.88\linewidth]{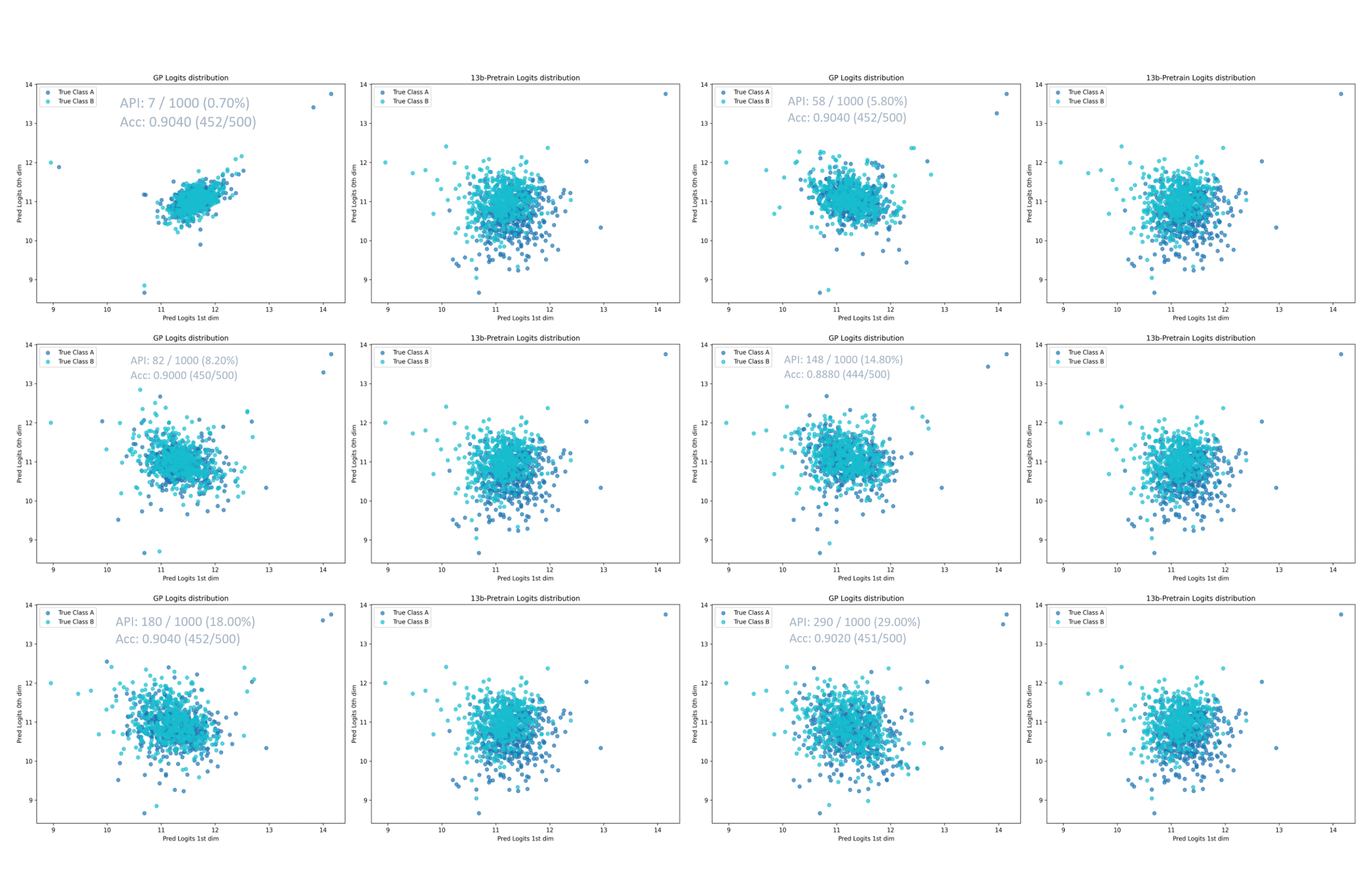}
    \caption{CoPA: across six API call budgets}
    \label{fig:copa_appendix}
  \end{subfigure}
  \vspace{0.2em}
  \begin{subfigure}[t]{\linewidth}
    \centering
    \includegraphics[width=0.88\linewidth]{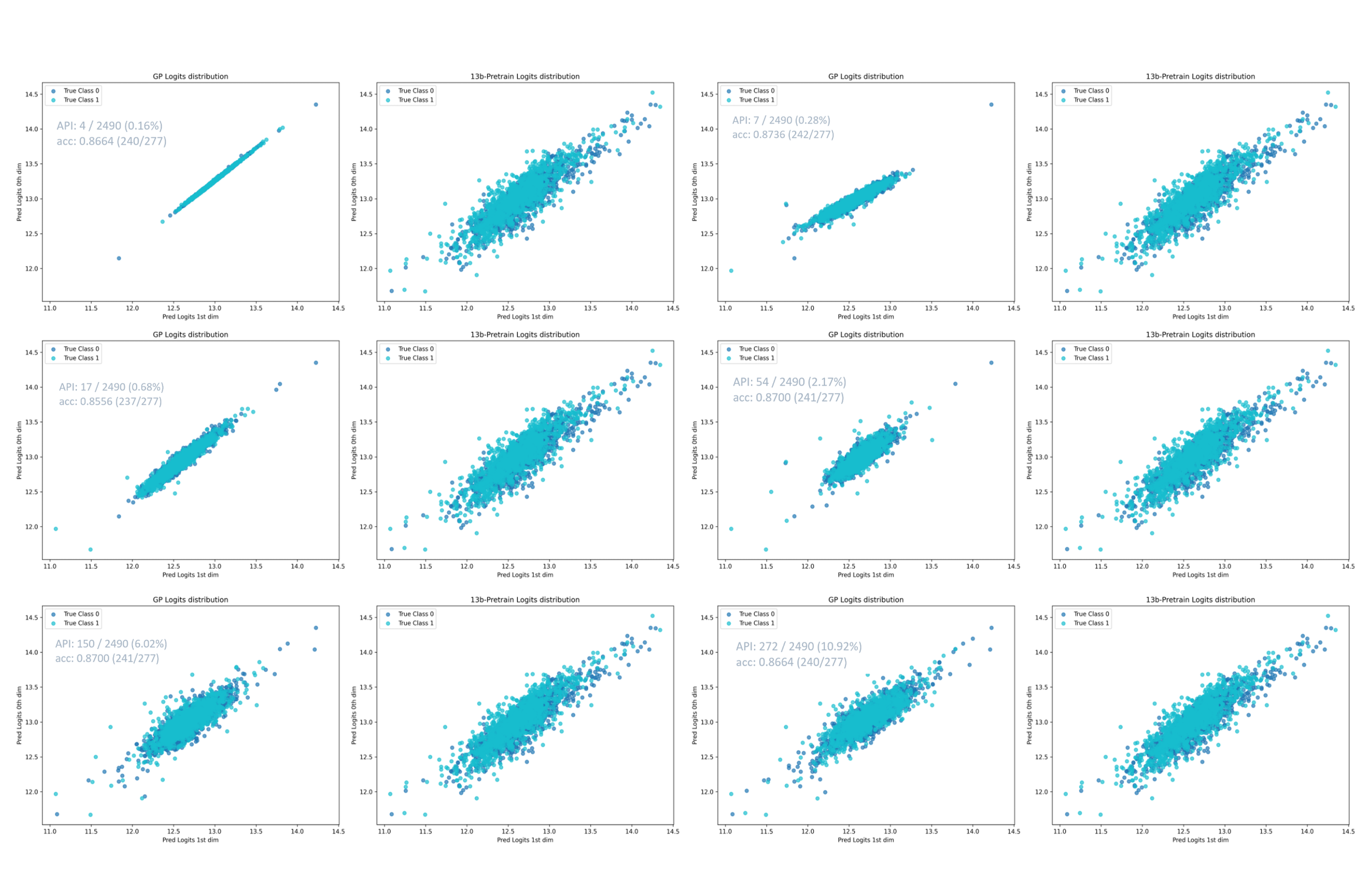}
    \caption{RTE: across six API call budgets}
    \label{fig:rte_appendix}
  \end{subfigure}
  \vspace{0.2em}
  \begin{subfigure}[t]{\linewidth}
    \centering
    \includegraphics[width=0.88\linewidth]{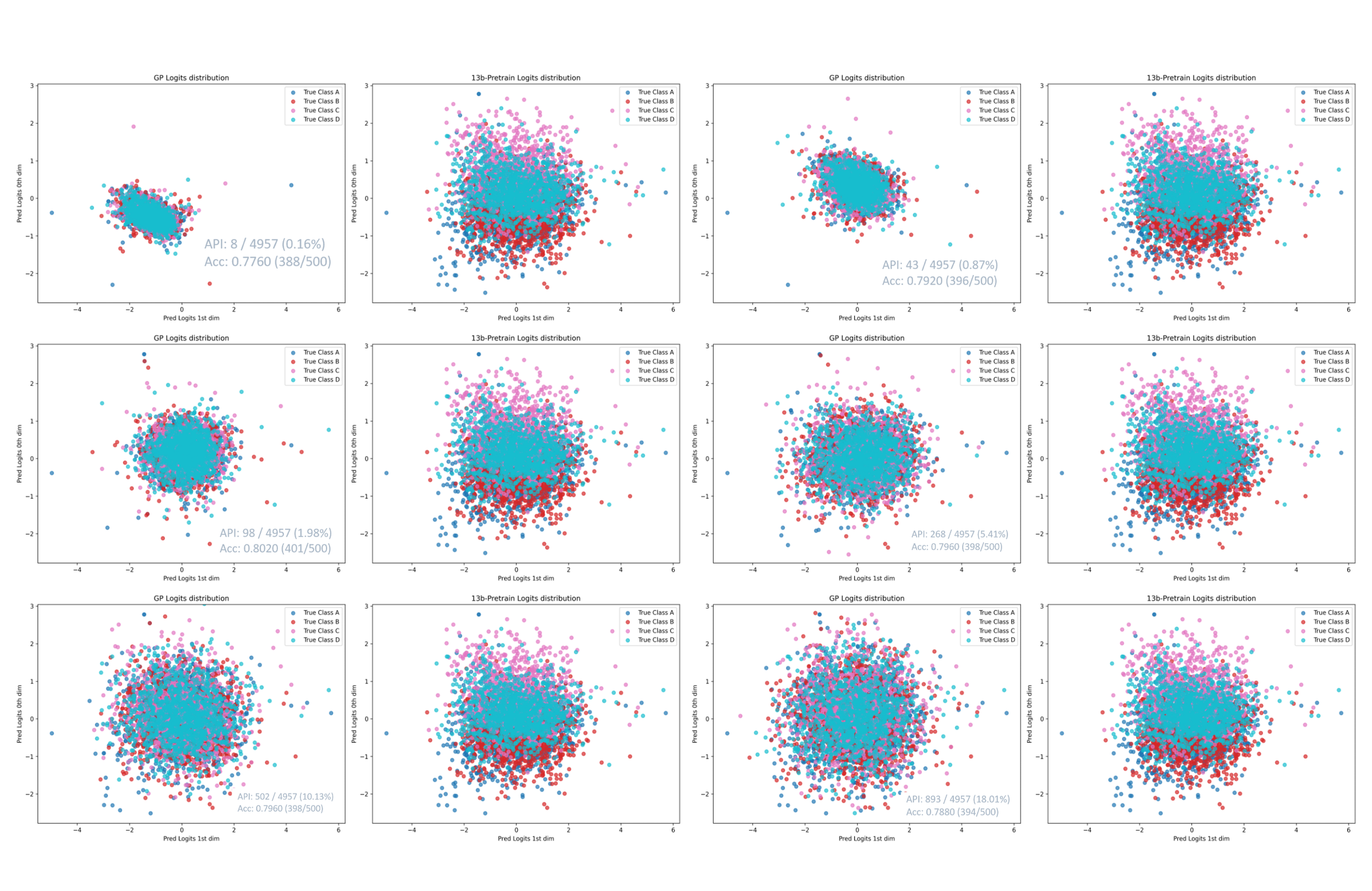}
    \caption{OBQA: across six API call budgets}
    \label{fig:obqa_appendix}
  \end{subfigure}
  \vspace{0.2em}
  \begin{subfigure}[t]{\linewidth}
    \centering
    \includegraphics[width=0.88\linewidth]{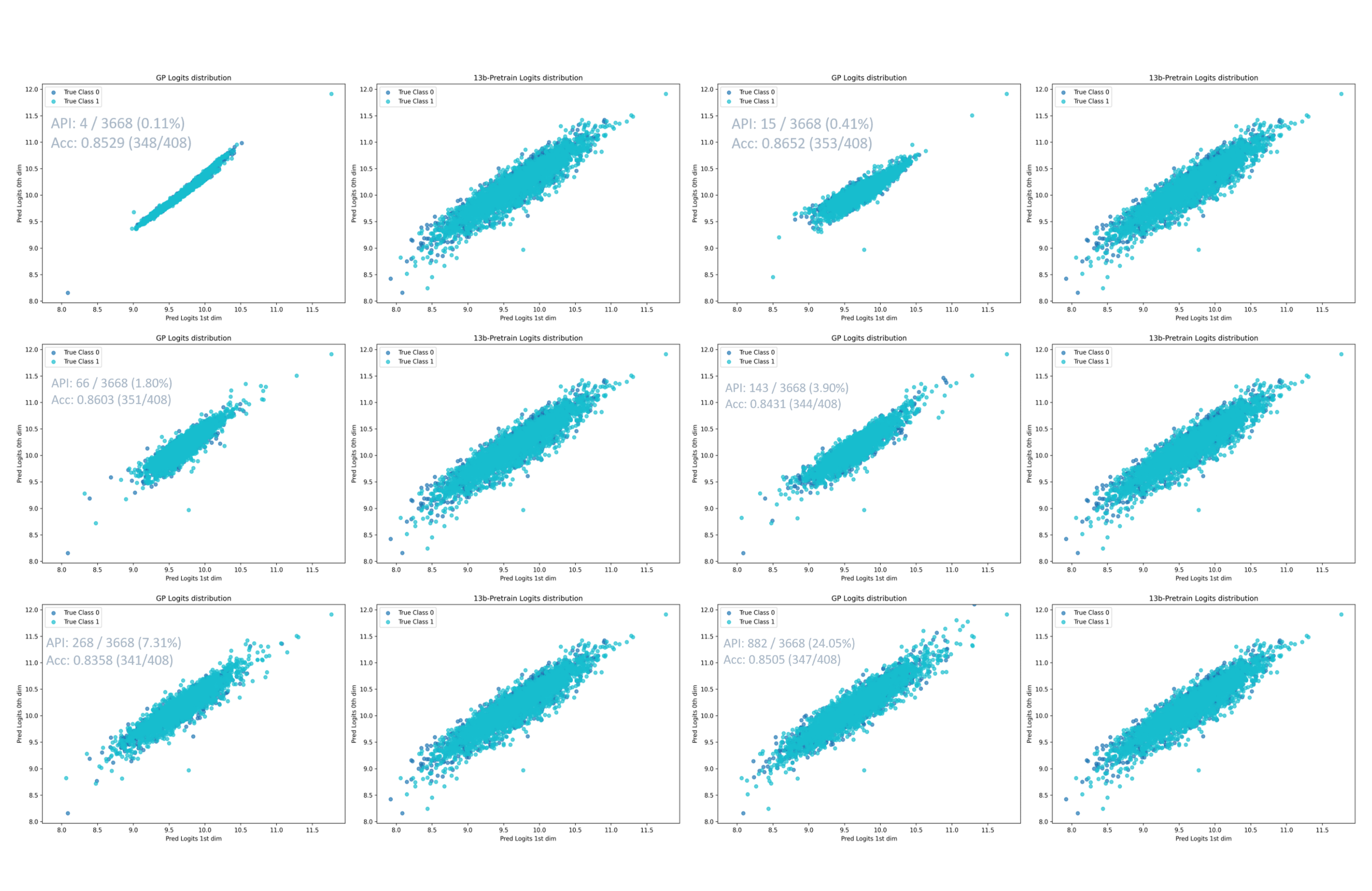}
    \caption{MRPC: across six API call budgets}
    \label{fig:mrpc_appendix}
  \end{subfigure}
  \caption{Supplementary distribution comparisons, complementing Figure~\ref{fig:logits_dist} from the main text.}
  \label{fig:logits_dist_appendix_all}
\end{figure}

While the main text discusses CoLA (Figure~\ref{fig:cola}) and ARC-C (Figure~\ref{fig:arcc}), this section presents the corresponding output logits distribution comparisons for the remaining four datasets: COPA, RTE, OBQA, and MRPC. The methodology for comparison remains consistent: for each dataset and varying API calls, we compare the logits distributions from GP models against those of the \textit{Llama2-13B} target model.

Figure~\ref{fig:logits_dist_appendix_all} displays these comparisons for the 4 additional datasets. As with the figures in main text (Figure~\ref{fig:logits_dist}), within each dataset's visualization, the left panel depicts the output logits distribution from the GP model, while the right panel shows the distribution from the large target model.

The visualizations for COPA (Figure~\ref{fig:copa_appendix}), RTE (Figure~\ref{fig:rte_appendix}), OBQA (Figure~\ref{fig:obqa_appendix}), and MRPC (Figure~\ref{fig:mrpc_appendix}) generally exhibit trends consistent with those detailed for CoLA and ARC-C in the main text. Notably, under severe API call limitations, the GP logit distributions tend to be more compressed. Conversely, with larger API call budgets, these distributions more closely mirror those of the 13B model. This consistency across diverse datasets and data regimes further underscores the robustness of our GP-filter method.


\section{Ablation Study Supplement}
\label{sec:appendix_ablations}

This section presents additional details and results from the ablation studies discussed in the main text, with a focus on evaluating the performance of our GP-filter method under extreme data scarcity, across challenging datasets, and under varying API call budgets.

\subsection{Performance under Extreme Data Scarcity}
\label{ablation-small}
The main text highlights that our GP-filter method maintains high performance even under conditions of extreme training data scarcity, where direct fine-tuning of proxy models (i.e., \textit{Llama2-7B}) can be significantly less effective. Table~\ref{tab:small-ex} presents the detailed experimental results that substantiate this claim. As indicated by the average scores, the GP-filter approach achieves an accuracy of 72.91\%, which is \textbf{6.31} percentage points higher than the 66.60\% achieved by full-precise fine-tuning \textit{Llama2-7B}. This strong performance is achieved with minimal API usage: the combined cost of GP model training and re-querying stages during proxy training averaged only \textbf{2.91\%} of the data in the scarcity experiments (see API Usage Percentage in Table~\ref{tab:small-ex}).

\begin{table}[h]
  \centering
  \renewcommand{\arraystretch}{1.5}
  \resizebox{\linewidth}{!}{
        \begin{tabular}{lcccccccccccc}
          \toprule
          \multirow{2}{*}{Model \& Method} & \multicolumn{12}{c}{\textbf{Accuracy (\%) $\uparrow$}} \\
          \cmidrule(lr){2-13}
                                       & AG-News & CoLA  & CoPA  & SST-2 & ARC-C & Cs-QA & OB-QA & MNLI  & QNLI  & RTE   & QQP   & Avg.  \\
          \midrule
          \textbf{7B Full Fine-tune}   & 92.88 & 74.40 & 61.40 & 81.31 & 43.81 & 32.76 & 53.40 & 87.61 & 63.41 & 54.15 & 87.42 & \textbf{66.60} \\
          Proxy-Tune                   & 92.63 & 74.88 & 78.40 & 80.85 & 56.19 & 49.96 & 57.60 & 87.55 & 74.70 & 57.76 & 87.45 & 72.54 \\
          \textbf{GP-filter}           & 92.68 & 73.73 & 78.80 & 83.14 & 56.52 & 49.71 & 57.20 & 88.06 & 76.66 & 58.12 & 87.40 & \textbf{72.91} \\
          \midrule
          Total Data Size              & 12,000 & 800 & 500 & 200 & 100 & 120 & 500 & 10,000 & 500 & 100 & 10,000 & - \\
          API Usage Amount             & 266    & 6   & 10  & 8  & 5   & 4   & 9   & 21     & 35  & 5   & 75     & - \\
          API Usage Percentage         & 2.22\% & 0.75\% & 2.00\% & 4.00\% & 5.00\% & 3.33\% & 1.80\% & 0.21\% & 7.00\% & 5.00\% & 0.75\% & 2.91\% \\
          \bottomrule
        \end{tabular}
  }
  \caption{Experiment on small subset training data. Performance is measured by Accuracy (\%).}
  \label{tab:small-ex}
\end{table}

\subsection{Performance on Challenging Datasets}
\label{ablation-hard}
As mentioned in the main text, our GP-filter method demonstrates a clear performance advantage over offline black-box tuning methods such as Proxy Tuning \cite{liu2024tuning}, particularly on inherently challenging datasets. This section presents a detailed comparative analysis on such tasks, with results summarized in Table~\ref{tab:hard-ex}.

To further assess robustness, we evaluate on three additional difficult datasets for large language models: the Microsoft Research Paraphrase Corpus (MRPC), Winograd Schema Challenge (WSC), and Adversarial Natural Language Inference (ANLI). Together with the three main datasets (ARC-C, Cs-QA, and RTE), this yields a suite of six challenging benchmarks. As shown in Table~\ref{tab:hard-ex}, our GP-filter method consistently and substantially outperforms Proxy-Tune across all tasks, achieving an average accuracy of 73.47\%, with \textbf{4.19} percentage point improvement over Proxy-Tune (69.28\%). Notably, this improvement is attained with only \textbf{1.99\%} average API usage of the large model $\mathcal{M}_l$, highlighting the efficiency of our method.

\begin{table}[h]
  \centering
  \renewcommand{\arraystretch}{1.4}
  \resizebox{\linewidth}{!}{
  \begin{tabular}{lccccccc}
    \toprule
    \multirow{2}{*}{Model \& Method} & \multicolumn{7}{c}{\textbf{Accuracy (\%) $\uparrow$ }} \\
    \cmidrule(lr){2-8}    & ARC-C & Cs-QA & RTE & MRPC & WSC & ANLI & Avg. \\
    \midrule
    7B Full Fine-tune    & 53.85 & 76.17 & 84.84 & 86.03 & 50.00 & 69.00 & 69.98 \\
    \textbf{Proxy-Tune}  & 57.53 & 75.35 & 82.67 & 84.07 & 47.67 & 68.40 & \textbf{69.28} \\
    \textbf{GP-filter}   & 61.20 & 78.95 & 87.73 & 87.50 & 54.65 & 70.80 & \textbf{73.47} \\
    \midrule
    GP API Usage         & 3.31\% & 1.95\% & 2.17\% & 2.37\% & 1.51\% & 0.64\% & 1.99\% \\
    \bottomrule
  \end{tabular}
  }
  \caption{Experiment on challenging datasets. Performance is measured by Accuracy (\%).}
  \label{tab:hard-ex}
\end{table}

The superior performance of our GP-filter method on challenging datasets stems from its fundamentally different approach to leveraging foundation model knowledge. In contrast, Proxy-Tune first trains a smaller proxy model in isolation and then combines its predictions with those of a zero-shot foundation model at inference time. On complex tasks, both components in Proxy-Tune can face limitations: the small proxy model is constrained by its capacity, while the large foundation model may be hampered by the inherent task difficulty or suboptimal prompting. Consequently, a simple inference-time amalgamation of their outputs, as employed by Proxy-Tune, often yields only marginal improvements or can even underperform a directly fine-tuned small model on these specific tasks.

In stark contrast, our GP-filter approach integrates the large model's insights more deeply and adaptively throughout the proxy model's training phase, addressing the aforementioned limitations:
\begin{itemize}
    \item \textbf{Continuous and Adaptive Guidance:} The GP surrogate model is trained to emulate the foundation model’s behavior, offering a rich, continuous, and adaptive guidance signal to the proxy model $\mathcal{M}^+_s$ during fine-tuning. This is substantially more effective than a static inference-time combination.
    \item \textbf{Improved Knowledge Internalization:} The GP model aims to capture the underlying “knowledge structure" or decision manifold of the large foundation model. By fine-tuning the proxy model $\mathcal{M}^+_s$ with this dynamically informed GP surrogate, $\mathcal{M}^+_s$ develops a more nuanced and robust task-specific representation, thereby elevating its overall performance ceiling and intrinsic capabilities.
\end{itemize}
The superior performance of our GP-guided fine-tuning on challenging datasets (Table~\ref{tab:hard-ex}) demonstrates the proxy model's enhanced ability to internalize sophisticated reasoning patterns. These empirical results underscore the importance of dynamically integrating knowledge from $\mathcal{M}_l$ during training, especially in scenarios where conventional offline proxy fine-tuning proves inadequate.

\subsection{Performance under Different API Usage}
\label{ablation-apiusage}

To further investigate the impact of API call volume on the performance of our GP surrogate model, we conducted a series of ablation studies. This section details our experiments on the COPA and ARC-C datasets, examining how varying the number of API calls used to train the GP surrogate influences its ability to guide the proxy model.

\begin{figure}[h]
  \centering
  \begin{subfigure}[t]{\linewidth}
    \centering
    \includegraphics[width=0.98\linewidth]{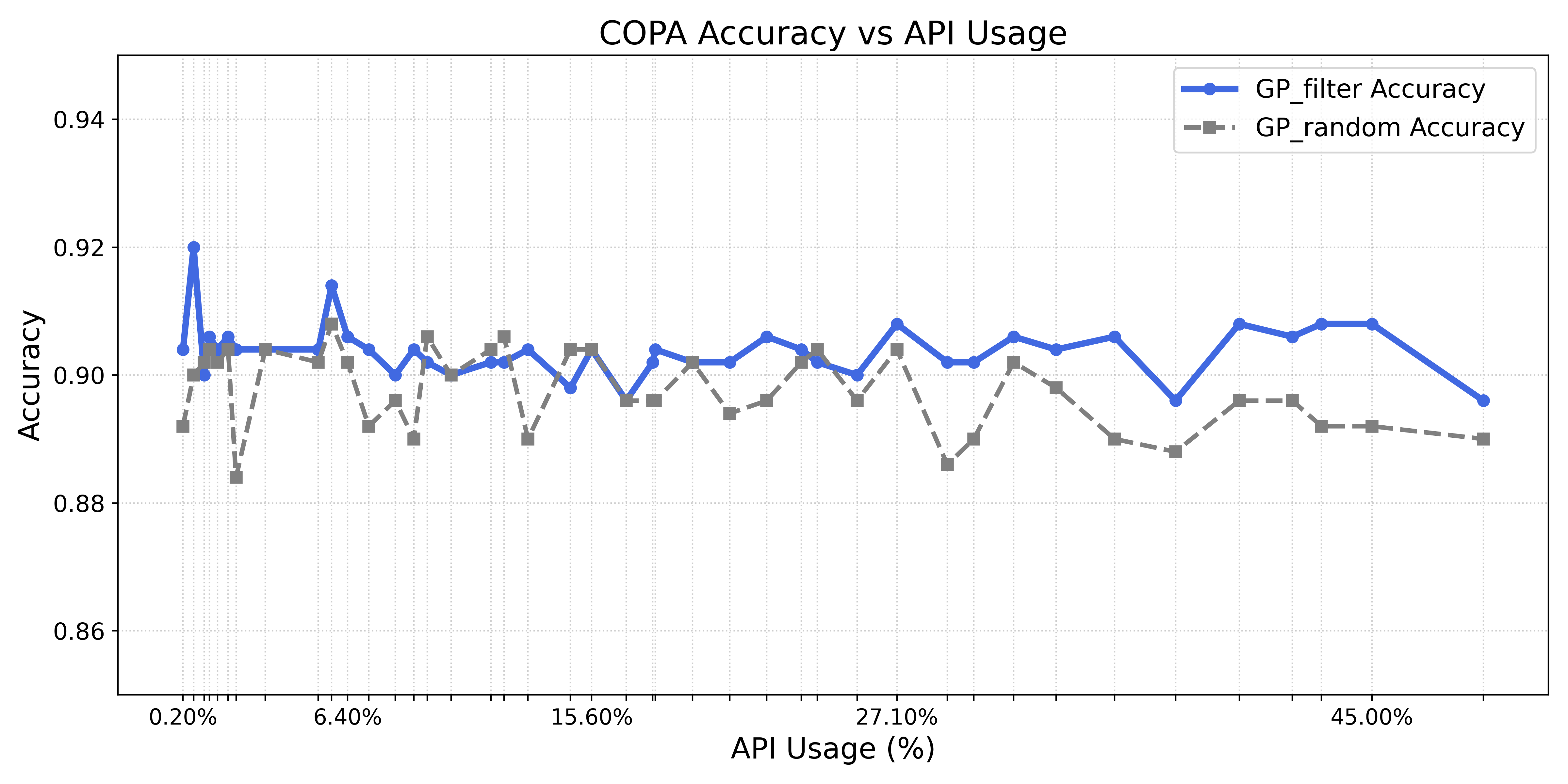}
    \caption{Accuracy on COPA with varying numbers of API calls}
    \label{fig:copa-accuracy}
  \end{subfigure}

  \vspace{1.5em}

  \begin{subfigure}[t]{\linewidth}
    \centering
    \includegraphics[width=0.98\linewidth]{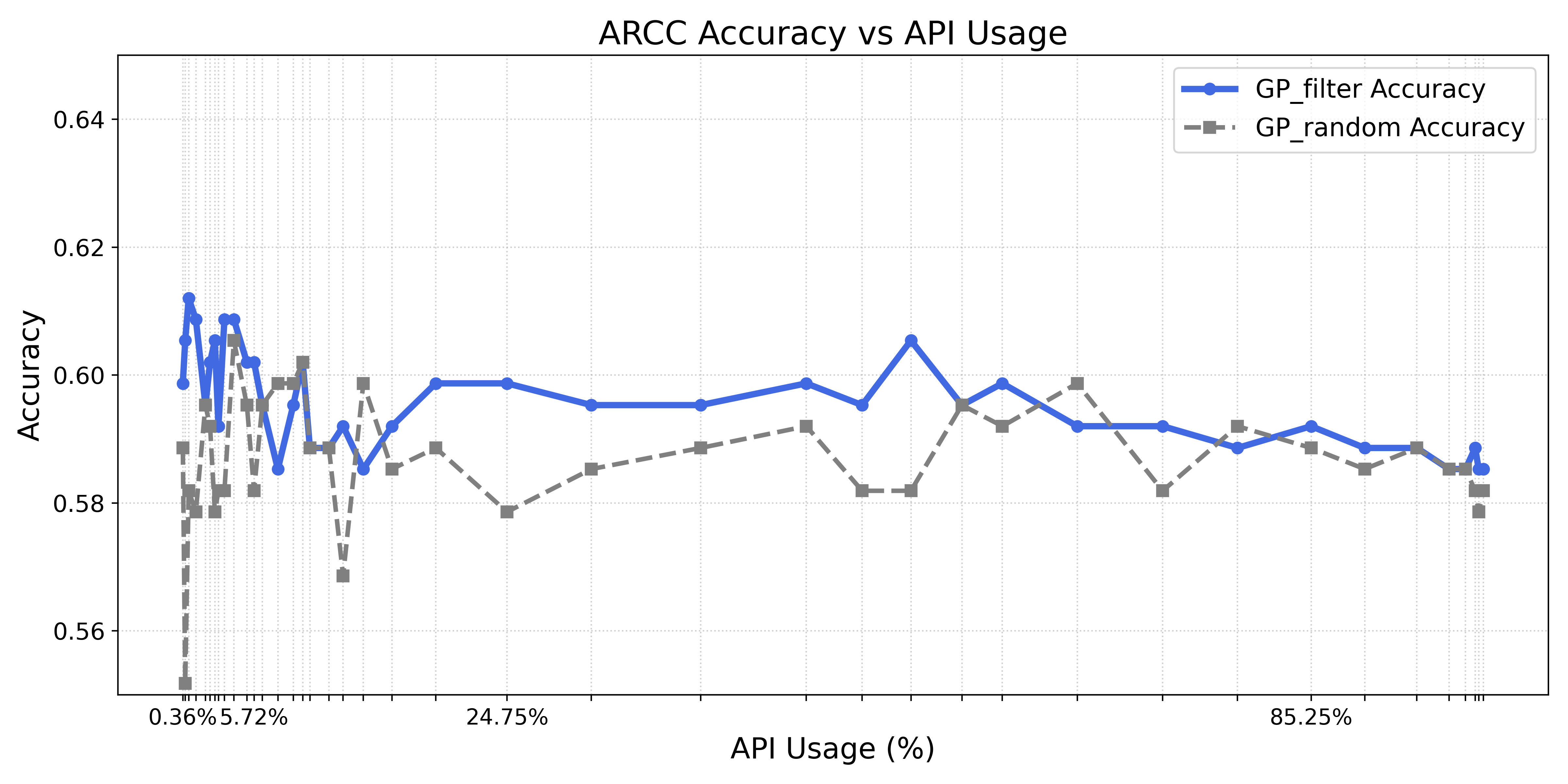}
    \caption{Accuracy on ARCC with varying numbers of API calls}
    \label{fig:arcc-accuracy}
  \end{subfigure}
  \caption{Accuracy comparison on COPA and ARCC with varying numbers of API calls used for GP model training.}
  \label{fig:accuracy-gp-copa-arcc}
\end{figure}

Our findings indicate that increasing the number of API calls does not necessarily lead to improved final accuracy. As illustrated in Figure \ref{fig:copa-accuracy} and Figure \ref{fig:arcc-accuracy}, the performance on both COPA and ARC-C datasets does not exhibit a monotonically increasing trend with more API calls. Although the GP surrogate's output distribution becomes progressively more refined to closely mimic the target model's (as detailed in Section~\ref{sec:distribution}), this refinement leads the overall proxy model performance to converge towards the CPT baseline, rather than achieving continuous improvement.

Interestingly, the figures highlight two key observations: First, peak accuracy is often achieved with a relatively small number of API calls. This aligns with our hypothesis that effective guidance from the GP surrogate hinges on capturing the overall structure of the target model's logits distribution, rather than requiring an exact replication of the logits themselves. Second, the consistent outperformance of our GP-filter method compared to the GP (random) approach underscores the effectiveness of our filter mechanism in facilitating this efficient knowledge transfer.

Furthermore, experiments on large-scale datasets such as MNLI and QQP (about 400,000 instances each) demonstrate that training the GP surrogate on a small filtered subset (e.g., 300 samples) can yield strong performance, even surpassing the CPT baseline. In contrast, using larger subsets (e.g., 10,000 samples) led to degraded performance and frequent GP training failures, including NaN predictions. These issues likely arise from two factors: (1) the $O(N^3)$ computational complexity of GP regression, which makes training on 10,000 samples impractical; and (2) increased risks of overfitting and numerical instability during kernel matrix inversion (especially with dense or collinear data points), which can result in NaN outputs. Our filtering strategy mitigates both challenges by selecting a diverse and informative subset, enabling stable and efficient GP training.

In summary, this ablation study underscores the effectiveness of a “less is more" strategy: a small, well-curated subset not only ensures stable and efficient learning, but also improves generalization, making GP-filter methods both scalable and effective in practice.


\section{Time and Memory Analysis of LogitMap Pair Construction and GP Training}
\label{sec:gp-ex}

To better assess the efficiency of our LogitMap pair construction and the subsequent Gaussian Process (GP) model training, we record both time and memory consumption under two distinct data selection strategies. Experiments are conducted on the QQP dataset~\cite{qqp} using models from the \textit{Mistral-7B} family~\cite{jiang2023mistral7b}. Specifically, as described in the “Empirical Data Selection" section in “Experiments":

\begin{itemize}
\item In the \textbf{random-based} setup, we randomly sample 5K training examples, following the upper bound suggested in the “Empirical Data Selection" section.
\item In the \textbf{filter-based} setup, we apply Algorithm~\ref{algo} to select approximately 3K informative examples (specifically, 2,825), also in line with the upper bound recommendation.
\end{itemize}

Execution time was recorded using Python's \texttt{time} library, while peak memory consumption was recorded via the \texttt{tracemalloc} library. For training GP model, we utilize \texttt{GaussianProcessRegressor} from \texttt{sklearn} library. We split the process into two phases.
\textbf{Phase 1:} Constructing LogitMap Pairs.
\textbf{Phase 2:} Training the GP model using the selected subset.

\begin{table}[h]
    \centering
    \begin{tabular}{l|cc|cc}
        \toprule
        \multirow{2}{*}{\textbf{Method}} & \multicolumn{2}{c|}{\textbf{Time (s)}} & \multicolumn{2}{c}{\textbf{Memory (MB)}} \\
        & Phase 1 & Phase 2 & Phase 1 & Phase 2 \\
        \midrule
        Random-based & 557.76 & 138.14 & 41.77 & 841.24 \\
        Filter-based & 1890.36 & 21.63 & 19.95 & 216.25 \\
        \bottomrule
    \end{tabular}
    \caption{Time and peak memory usage for constructing LogitMap Pairs (Phase 1) and training the GP model (Phase 2) under two data selection strategies on the QQP dataset.}
    \label{tab:gp-ex}
\end{table}

As shown in Table~\ref{tab:gp-ex}, both phases are highly efficient in terms of time and memory consumption. Notably, even in the computationally intensive \textbf{filter-based} setup, the total time required is approximately 1,912 seconds, or just over \textbf{0.53 hours}. In contrast, fine-tuning the small proxy model requires approximately \textbf{16 hours} on 6 RTX 4090 GPUs for a standard two-epoch training, highlighting the negligible overhead of our data selection and GP-fitting pipeline.

From a memory perspective, both strategies maintain a remarkably low footprint. The peak memory usage stays well below \textbf{1 GB} in all phases, with the filter-based strategy consuming only 216.25 MB during GP training—further highlighting the lightweight nature of our method. These results demonstrate that our approach not only scales effectively to large datasets but also offers a practical, resource-efficient alternative to conventional proxy tuning.


\section{Real-World Black-box Tuning Experiment}
\label{sec:realblackbox}

We accessed the Qwen-Plus model via API calls as follows:  

\vspace{1em}

\begin{lstlisting}[language=Python]
completion = client.chat.completions.create(
    model="qwen-plus",
    messages=[
        {"role": "system", "content": "You are a helpful assistant."},
        {"role": "user", "content": prompt}
    ],
    logprobs=True,
    top_logprobs=5,
    max_tokens=512
)
lp = completion.choices[0].logprobs.content[0].top_logprobs
\end{lstlisting}

\vspace{1em}

In realistic black-box settings, LLM providers (e.g., ChatGPT, Qwen) typically offer log-probability access limited to only the top-5 most likely tokens. For our GP-filter tuning, we leverage this constraint by extracting the top-5 token IDs and their associated log-probabilities from the black-box model's response. These token IDs are then aligned with the white-box proxy model's vocabulary to enable logit-level supervision.

Despite this restriction, we find that careful prompt design can ensure the critical tokens are consistently included in the top-5 set. For instance, when using classification-style prompts (e.g., yes / no questions), the relevant tokens (such as “yes" and “no") almost always appear among the top-5 predictions. In generative settings, we assume the black-box model's top-ranked token IDs to guide the white-box model's fine-tuning, ensuring stylistic and semantic alignment between the two. This strategy allows our method to operate effectively within realistic API constraints, while still achieving precise logit-level guidance.

We evaluate on the RTE (Recognizing Textual Entailment) dataset~\cite{dagan2005pascal} using Qwen3-series models (1.7B, 4B, 8B, 14B)~\cite{yang2025qwen3technicalreport} as white-box proxies. Results are summarized in Table~\ref{tab:realblackbox}.

\begin{table}[h]
  \centering
  \resizebox{\linewidth}{!}{%
  \begin{tabular}{lcccc}
    \toprule
    \textbf{Method}         & \textbf{1.7B} & \textbf{4B} & \textbf{8B} & \textbf{14B} \\
    \midrule
    Qwen3-Pretrain          & 66.18\%       & 84.73\%     & 85.82\%     & 83.64\%      \\ [1ex]
    Qwen3-GP-tuned          & 80.73\%       & 86.55\%     & 87.64\%     & 92.36\%      \\ [2ex]
    Qwen-Plus (origin)      & \multicolumn{4}{c}{90.91\%}                              \\ [1ex]
    Qwen-Plus (GP-filter)   & 91.27\%       & 91.27\% & \textbf{92.36\%} & \textbf{93.45\%}   \\ [2ex]
    Few-shot (1-shot)       & \multicolumn{4}{c}{90.91\%}                              \\ [1ex]
    Few-shot (5-shot)       & \multicolumn{4}{c}{91.27\%}                              \\ [1ex]
    Few-shot (10-shot)      & \multicolumn{4}{c}{91.64\%}                              \\ 
    \bottomrule
  \end{tabular}
  }
  \caption{Validation accuracies on RTE under GP-filter black-box tuning. Notes: two RTE validation examples were filtered by the provider's safety mechanism, resulting in a total of 275 examples.}
  \label{tab:realblackbox}
\end{table}

Table~\ref{tab:realblackbox} reports the RTE validation accuracies under various inference and tuning settings. The row “Qwen3-Pretrain" shows zero‐shot performance using the original pretrained weights of the four Qwen3 proxy models (1.7B–14B). “Qwen3‐GP–tuned" gives the accuracies after fine‐tuning those proxies with our GP‐filter method. “Qwen-Plus (origin)" is direct zero‐shot inference with the black-box Qwen-Plus model, while “Qwen-Plus (GP-filter)" performs inference by ensembling the logits from Qwen3 proxies and the Qwen-Plus model (see Equation~\ref{eq:inference}). “Few-shot" refers to the performance of the black-box Qwen-Plus model under few-shot inference.

It is immediately clear that GP-filter yields consistent improvements: for example, the Qwen-Plus improves from 90.91\% to 93.45\% when ensembled with GP-tuned Qwen3-14B proxies. Crucially, we only issued 31 API calls to Qwen-Plus to train the Gaussian Process surrogate (total 2,490 training samples), demonstrating exceptional effectiveness and cost‐efficiency.

By contrast, static few-shot prompting requires much larger input contexts (slower inference) and delivers no substantial accuracy gains (1-shot: 90.91\%, 5-shot: 91.27\%, 10-shot: 91.64\%). We attribute this to few-shot examples teaching only formatting conventions rather than the model’s internal reasoning. In contrast, GP-filter integrates black-box model guidance directly into proxy training, effectively transferring downstream dataset knowledge. This advantage is especially pronounced on harder examples: further increasing the number of few-shot examples fails to improve accuracy, whereas our GP-filter method continues to yield gains (see Appendix~\ref{ablation-hard}).

\end{document}